\documentclass[runningheads]{llncs}
\usepackage[width=122mm,left=12mm,paperwidth=146mm,height=193mm,top=12mm,paperheight=217mm]{geometry}
\usepackage{graphicx}
\usepackage{amsmath,amssymb} % define this before the line numbering.
\usepackage{color}
\usepackage[hidelinks]{hyperref}
\usepackage{rotating}
\usepackage{threeparttable}
\usepackage{array}
\newcolumntype{C}[1]{>{\centering\let\newline\\\arraybackslash\hspace{0pt}}m{#1}}

\begin{document}
\title{Deep Learning-based Image Super-Resolution Considering Quantitative and Perceptual Quality} 
% Replace with your title

\titlerunning{Super-Resolution Considering Quantitative and Perceptual Quality}
% Replace with a meaningful short version of your title
%
\author{Jun-Ho Choi \and Jun-Hyuk Kim \and Manri Cheon \and Jong-Seok Lee}
%
%Please write out author names in full in the paper, i.e. full given and family names. 
%If any authors have names that can be parsed into FirstName LastName in multiple ways, please include the correct parsing, in a comment to the volume editors:
%\index{Lastnames, Firstnames}
%(Do not uncomment it, because you may introduce extra index items if you do that, we will use scripts for introducing index entries...)
\authorrunning{Choi et al.}
% Replace with shorter version of the author list. If there are more authors than fits a line, please use A. Author et al.
%

\institute{School of Integrated Technology, Yonsei University, Korea \\
	\email{\{idearibosome, junhyuk.kim, manri.cheon, jong-seok.lee\}@yonsei.ac.kr}}
\maketitle              % typeset the header of the contribution
\begin{abstract}
Recently, it has been shown that in super-resolution, there exists a tradeoff relationship between the quantitative and perceptual quality of super-resolved images, which correspond to the similarity to the ground-truth images and the naturalness, respectively.
In this paper, we propose a novel super-resolution method that can improve the perceptual quality of the upscaled images while preserving the conventional quantitative performance.
The proposed method employs a deep network for multi-pass upscaling in company with a discriminator network and two quantitative score predictor networks.
Experimental results demonstrate that the proposed method achieves a good balance of the quantitative and perceptual quality, showing more satisfactory results than existing methods.

\end{abstract}

\section{Introduction}

Single-image super-resolution, which is a task to increase the spatial resolution of low-resolution images, has been widely studied in recent decades.
One of the simple solutions for the task is to employ interpolation methods such as nearest-neighbor and bicubic upsampling.
However, their outputs are largely blurry because fine details of the images cannot be recovered.
Therefore, many researchers have investigated how to effectively restore high-frequency details.
Nevertheless, it is still highly challenging due to the lack of information in the low-resolution images, i.e., an ill-posed problem \cite{ledig2017photo}.

Until the mid-2010s, feature extraction-based methods have been proposed, including sparse coding \cite{yang2011multitask}, neighbor embedding \cite{li2014single}, and Bayes forest \cite{salvador2015naive}.
After that, the emergence of deep learning for visual representation \cite{guo2016deep}, which is triggered by an image classification challenge (i.e., ImageNet) \cite{krizhevsky2012imagenet}, has also flowed into the field of super-resolution \cite{yang2018deep}.
For instance, the super-resolution convolutional neural network (SRCNN) model proposed by Dong \textit{et al.} \cite{dong2014learning} introduced convolutional layers and showed better performance than the previous methods.

\begin{figure}[t]
	\centering
	\begin{minipage}[b]{0.24\linewidth}
		\centering
		\centerline{\includegraphics[width=0.99\linewidth]{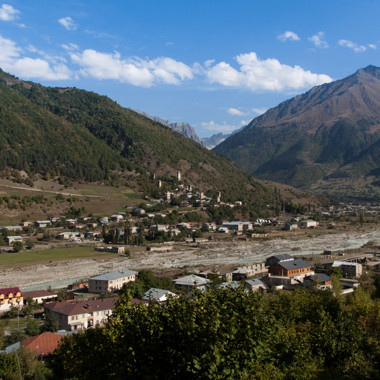}}
		\centerline{(a)}
	\end{minipage}
	\begin{minipage}[b]{0.24\linewidth}
		\centering
		\centerline{\includegraphics[width=0.99\linewidth]{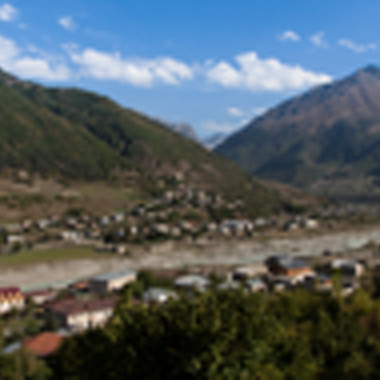}}
		\centerline{(b)}
	\end{minipage}
	\begin{minipage}[b]{0.24\linewidth}
		\centering
		\centerline{\includegraphics[width=0.99\linewidth]{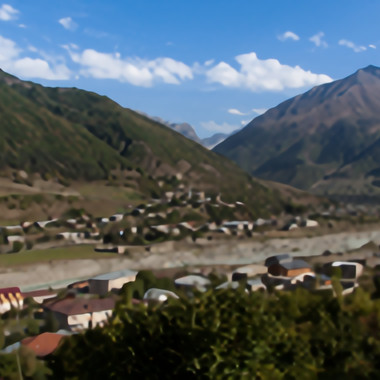}}
		\centerline{(c)}
	\end{minipage}
	\begin{minipage}[b]{0.24\linewidth}
		\centering
		\centerline{\includegraphics[width=0.99\linewidth]{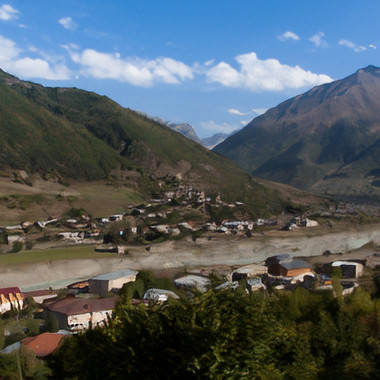}}
		\centerline{(d)}
	\end{minipage}
	\caption{Example results obtained for an image of the PIRM dataset \cite{blau20182018}. (a) Ground-truth (b) Upscaled by bicubic interpolation (c) Upscaled without perceptual consideration (d) Upscaled with perceptual consideration}
	\label{fig:example_result}
\end{figure}

To build a deep learning-based super-resolution model, it is required to define loss functions that are the objectives of the model to be trained.
Loss functions measuring pixel-by-pixel differences of the ground-truth and upscaled images are frequently considered, including mean squared error and mean absolute error \cite{yang2018deep}.
They mainly aim at guaranteeing quantitative conditions of the obtained images, which can be evaluated by quantitative quality measures such as peak signal-to-noise ratio (PSNR), root mean squared error (RMSE), and structural similarity (SSIM) \cite{wang2004image}.
\figurename~\ref{fig:example_result}~(c) shows an example image generated by a deep learning-based super-resolution model, enhanced upscaling super-resolution (EUSR) \cite{kim2018deep}, from the downscaled version of \figurename~\ref{fig:example_result}~(a).
Compared to the image upscaled by bicubic interpolation shown in \figurename~\ref{fig:example_result}~(b), the image generated by the deep learning-based method follows the overall appearance of the original image with sharper boundaries of the objects and scenery.

Although existing methods based on minimizing pixel-by-pixel differences achieve great performance in a quantitative viewpoint, they do not ensure \textit{naturalness} of the output images.
For example, fine details of trees and houses are not sufficiently recovered in \figurename~\ref{fig:example_result}~(c).
To improve the naturalness of the images, two approaches have been proposed in the literature: using generative adversarial networks (GANs) \cite{goodfellow2014generative} and employing intermediate features of the common image classification network models.
For example, Ledig \textit{et al.} \cite{ledig2017photo} proposed a super-resolution model named SRGAN, which employs a discriminator network and trains the model to minimize differences of the intermediate features of VGG19 \cite{simonyan2014very} when the ground-truth and upscaled images are inputted.
It is known that these methods enhance perceptual performance significantly \cite{blau2017perception}.
Here, the perceptual performance can be measured by the metrics for visual quality assessment such as blind/referenceless image spatial quality evaluator (BRISQUE) \cite{mittal2012no} and naturalness image quality evaluator (NIQE) \cite{mittal2013making}.

However, two issues still remain unresolved in these approaches.
First, although these approaches improve naturalness of the images, perceptual quality is only indirectly considered and thus the improvement may be limited.
The network models for extracting intermediate features are for image classification tasks, thus forcing the features to be similar does not guarantee perceptually improved results.
In addition, it is possible that the discriminator network learns the criteria that can differentiate generated images from the real ones but are not related to the perceptual aspects.
For instance, when the trained discriminator relies on just finding high-frequency components, the super-resolution model may add some unexpected textures in low-frequency regions such as ground and sky.

Second, these approaches tend to sacrifice a large amount of the quantitative quality.
For example, the SRGAN-based models achieve better perceptual performance than the other models in terms of BRISQUE and NIQE, but they record worse quantitative quality, showing larger RMSE values \cite{blau2017perception}.
Since the primary objective of the super-resolution task is to make the upscaled images identical to the ground-truth high-resolution images, it is necessary to properly regularize the upscaling modules to keep balance of the quantitative and qualitative quality.

In this paper, we propose a novel super-resolution method named ``Four-pass perceptual super-resolution with enhanced upscaling (4PP-EUSR),'' which is based on the recently proposed EUSR model \cite{kim2018deep}.
Our model aims at resolving the aforementioned issues via two innovative ways.
First, our model employs so-called ``multi-pass upscaling'' during the training phase, where multiple upscaled images produced by passing the given low-resolution image through the multiple upscaling paths in our model are used in order to consider various possible characteristics of upscaled images.
Second, we employ qualitative score predictors, which directly evaluate the aesthetic and subjective quality scores of the upscaled images.
This architecture ensures high perceptual quality with preserving the quantitative performance of the upscaled images, as exemplified in \figurename~\ref{fig:example_result}~(d).

The rest of the paper is organized as follows.
First, we provide a brief review of the related work in Section~\ref{sec:related_work}.
Then, an overview of the proposed method is given in Section~\ref{sec:method}, including the base deep learning model, multi-pass upscaling for training, structure of the discriminator, and structures of the qualitative score predictors.
We explain training procedures of our model with the employed loss functions in Section~\ref{sec:training_details}.
In-depth experimental analysis of our results is shown in Section~\ref{sec:results}.
Finally, we conclude our work in Section~\ref{sec:conclusion}.

\section{Related work}
\label{sec:related_work}

In this section, we review the related work of deep learning-based super-resolution in two branches: super-resolution models without and with consideration of naturalness.

\subsection{Deep learning-based super-resolution}

One of the earliest super-resolution models based on deep learning is SRCNN, which was proposed by Dong \textit{et al.} \cite{dong2014learning}.
The model takes an image upscaled by the bicubic interpolation and enhances it via two convolutional layers.
Kim \textit{et al.} proposed the very deep super-resolution (VDSR) model \cite{kim2016accurate}, which consists of 20 convolutional layers.
In recent days, residual blocks having shortcut connections \cite{he2016deep} are commonly used in the super-resolution models.
For example, Ledig \textit{et al.} \cite{ledig2017photo} proposed a model named SRResNet, which contains 16 residual blocks with batch normalization \cite{ioffe2015batch} and parametric ReLU activation \cite{he2015delving}.
Lim \textit{et al.} \cite{lim2017enhanced} developed two super-resolution models for the NTIRE 2017 single-image super-resolution challenge \cite{timofte2017ntire}: the enhanced deep super-resolution (EDSR) model for single-scale super-resolution and the multi-scale deep super-resolution (MDSR) model for multi-scale super-resolution.
They found that removing batch normalization and blending outputs generated from geometrically transformed inputs help improving the overall quantitative quality.
Recently, Kim and Lee \cite{kim2018deep} suggested a multi-scale super-resolution method named EUSR, which consists of so-called ``enhanced upscaling modules'' and performed well in the NTIRE 2018 single-image super-resolution challenge \cite{timofte2018ntire}.
Zhang \textit{et al.} \cite{zhang2018residual} proposed a super-resolution model based on residual dense network (RDN), which extends the residual network to have densely-connected layers.
Zhang \textit{et al.} \cite{zhang2018image} proposed a residual channel attention networks (RCAN), which brings an attention mechanism into the super-resolution task and achieves better quantitative performance than EDSR.

\subsection{Super-resolution considering naturalness}

Along with ensuring high quantitative quality in terms of PSNR, RMSE, or SSIM, naturalness of the upscaled images, which can be measured by quality metrics such as BRISQUE and NIQE, has been also considered in some studies.
There exist two common approaches: employing GANs \cite{goodfellow2014generative} and employing image classifiers.
In the former approach, the discriminator network tries to distinguish the ground-truth images from the upscaled images and the super-resolution model is trained to fool the discriminator so that it cannot distinguish the upscaled images properly.
When an image classifier is used, the super-resolution model is trained to minimize the difference of the features obtained at the intermediate layers of the classifier for the ground-truth and upscaled images.
For example, Johnson \textit{et al.} \cite{johnson2016perceptual} used the trained VGG16 network to extract the intermediate features and regarded the squared Euclidean distance between them as the loss function.
Ledig \textit{et al.} \cite{ledig2017photo} employed an adversarial network and differences of the features obtained from the trained VGG19 network for calculating losses of their super-resolution model (i.e., SRResNet), which is named as SRGAN.
Mechrez \textit{et al.} \cite{mechrez2018learning} defined the so-called ``contextual loss,'' which compares the statistical distribution of the intermediate features obtained from the trained VGG19 model, to train their super-resolution model.
These models focus on ensuring naturalness of the upscaled images but tend to sacrifice a large amount of the quantitative quality \cite{blau2017perception}.

\section{Overview of the proposed method}
\label{sec:method}

\begin{figure}[t]
	\centering
	\includegraphics[width=0.6\linewidth]{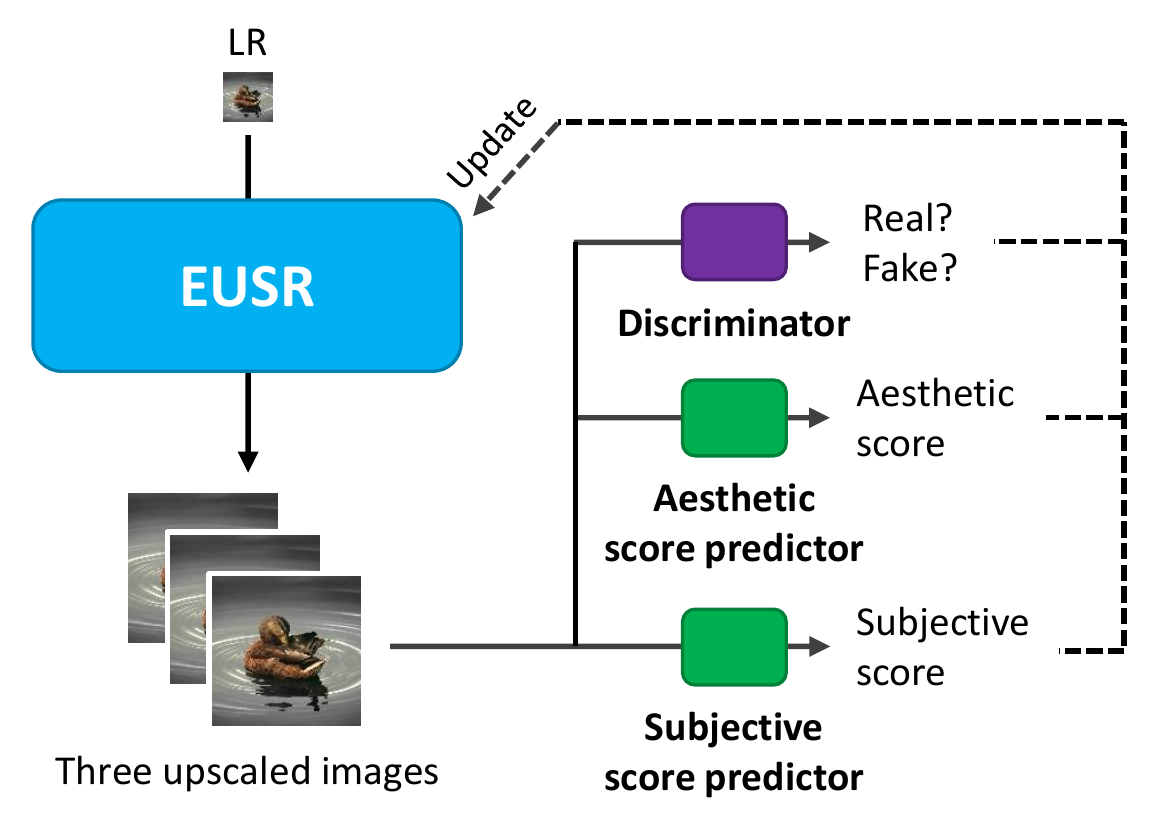}
	\caption{Overview of the proposed method. First, our super-resolution model (Section~\ref{sec:eusr}) generates three upscaled images via multi-pass upscaling (Section~\ref{sec:multipass_upscaling}). The discriminator tries to differentiate the upscaled images from the ground-truth (Section~\ref{sec:discriminator}). The two qualitative score predictors measure the aesthetic and subjective quality scores, respectively (Section~\ref{sec:qualitative_score_predictors}). The outputs of the discriminator and the score predictors are used to update the super-resolution model.}
	\label{fig:method_procedure}
\end{figure}

The architecture of the proposed method can be disassembled into four components (\figurename~\ref{fig:method_procedure}): a multi-scale upscaling model, employing the model in a multi-pass manner, a discriminator, and qualitative score predictors.

\subsection{Enhanced upscaling super-resolution}
\label{sec:eusr}

\begin{figure*}[t]
	\centering
	\includegraphics[width=0.98\linewidth]{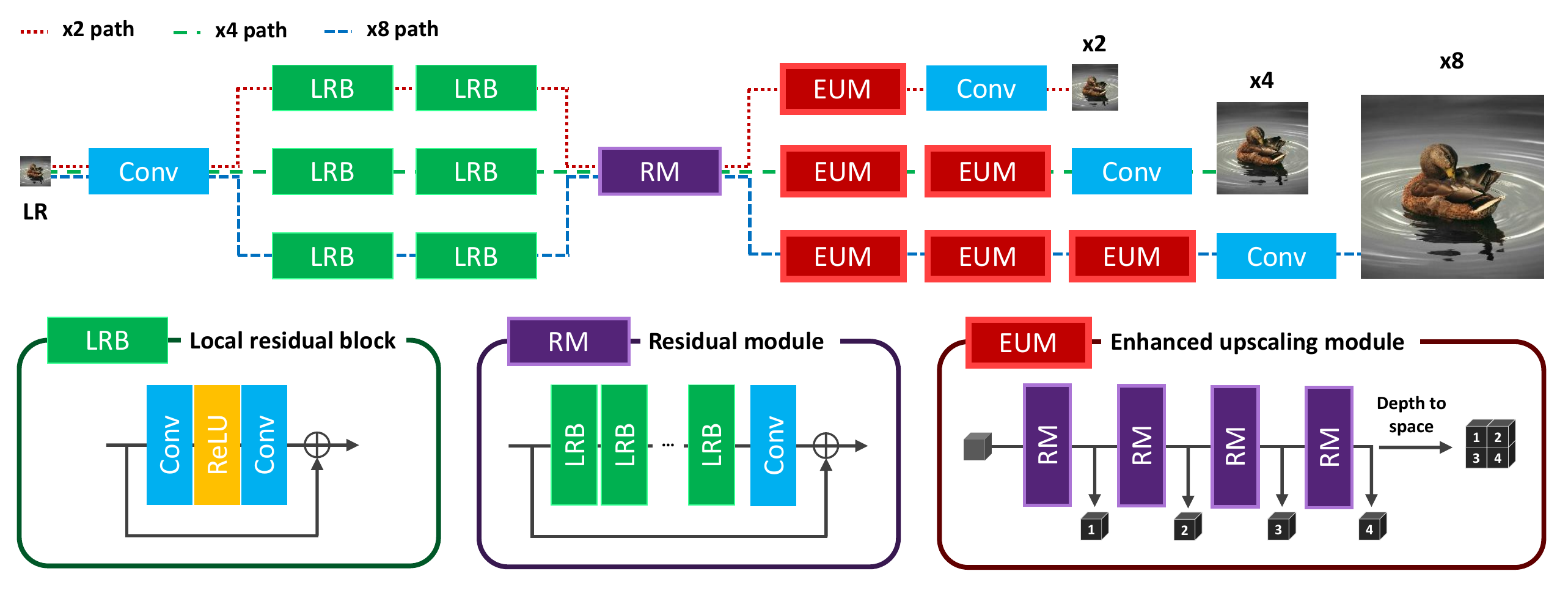}
	\caption{Structure of the EUSR model \cite{kim2018deep}.}
	\label{fig:eusr_structure}
\end{figure*}

The basic structure of our model is from the EUSR model \cite{kim2018deep}, which is shown in \figurename~\ref{fig:eusr_structure}.
It mainly consists of three parts: scale-aware feature extraction, shared feature extraction, and enhanced upscaling.
First, the scale-aware feature extraction part extracts low-level features from the input image by using so-called ``local residual blocks.''
Then, a residual module in the shared feature extraction part, which consists of local residual blocks and a convolutional layer, extracts higher-level features regardless of the scale factor.
Finally, the proceeded features are upscaled via ``enhanced upscaling modules,'' where each module increases the spatial resolution of the input by a factor of 2.
Thus, the $\times$2, $\times$4, and $\times$8 upscaling paths have one, two, and three enhanced upscaling modules, respectively.
The configurable parameters of the EUSR model are the number of output channels of the first convolutional layer, the number of local residual blocks in the shared feature extraction part, and the number of local residual blocks in the enhanced upscaling modules.
We consider EUSR as our base upscaling model because it is one of the state-of-the-art approaches supporting multi-scale super-resolution, which enables generating multiple upscaled images from a single model.

\subsection{Multi-pass upscaling}
\label{sec:multipass_upscaling}

\begin{figure}[t]
	\centering
	\includegraphics[width=0.6\linewidth]{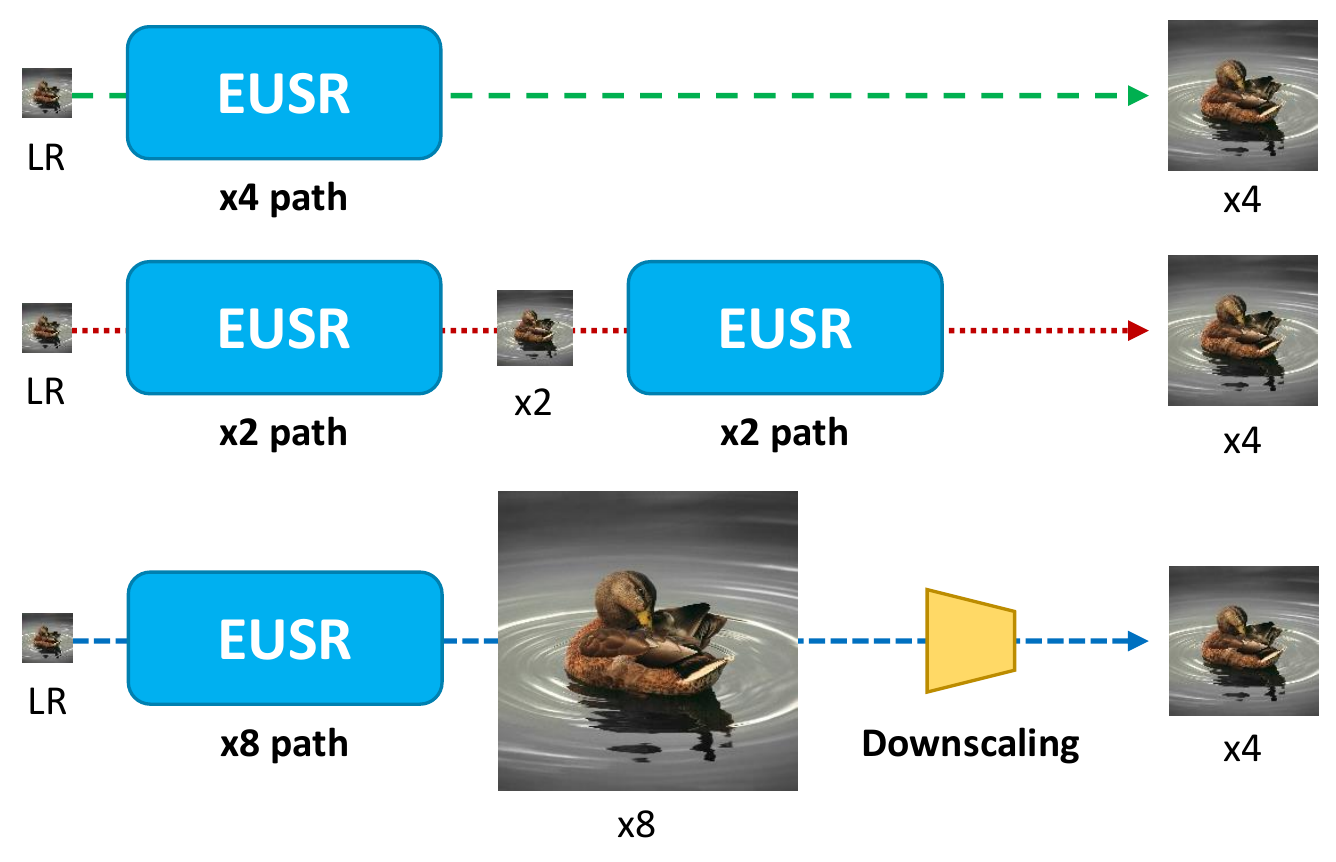}
	\caption{Multi-pass upscaling process, which produces three upscaled images by a factor of 4 from a shared pre-trained EUSR model.}
	\label{fig:4pp_eusr_structure}
\end{figure}

The original EUSR model supports multi-scale super-resolution by factors of 2, 4, and 8.
During the training phase, our model utilizes all these upscaling paths to produce three output images, where we make the output images have the same upscaling factor of 4 for a given image as follows (\figurename~\ref{fig:4pp_eusr_structure}).
The first one is directly generated from the $\times$4 path.
The second one is generated by passing the given image through the $\times$2 path two times.
The third one is generated via bicubic downscaling of the image obtained from the $\times$8 path by a factor of 2.
Thus, the model is employed four times for each input image.

The original purpose of multi-scale models such as MDSR \cite{lim2017enhanced} and EUSR \cite{kim2018deep} is to support variable scaling factors on a single model.
On the other hand, our multi-pass upscaling extends it with a different objective, which is to improve the quality of the upscaled images for a fixed scaling factor.
Since all three images obtained from different upscaling paths are used for training, the model has to learn reducing artifacts that may occur during direct upscaling via the $\times$4 path, two-pass upscaling via the $\times$2 path, and upscaling via the $\times$8 path and downscaling.
This prevents the model to overfit towards specific patterns, thus it enables the model to handle various upscaling scenarios.
%Note that although the multi-pass upscaling looks similar to the geometric self-ensemble strategy that is used in various super-resolution methods (e.g., EDSR \cite{lim2017enhanced} and EUSR \cite{kim2018deep}), 

\subsection{Discriminator network}
\label{sec:discriminator}

\begin{figure*}[t]
	\centering
	\includegraphics[width=0.98\linewidth]{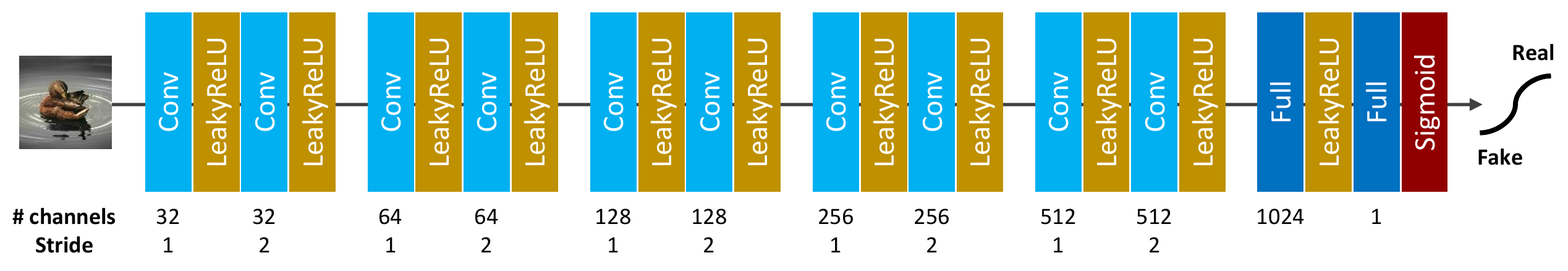}
	\caption{Structure of the discriminator network.}
	\label{fig:discriminator_structure}
\end{figure*}

Our method employs a discriminator network during the training phase, which is designed to distinguish generated images from the ground-truth images.
While the discriminator tries to do its best for identifying the upscaled images, the super-resolution model is trained to make the discriminator difficult to differentiate them from the ground-truth images.
This helps our upscaling model generating more natural images \cite{ledig2017photo,mechrez2018learning}.
Inspired by SRGAN \cite{ledig2017photo}, our discriminator network consists of several convolutional layers followed by LeakyReLU activations with $\alpha = 0.2$ and two fully-connected layers, as shown in \figurename~\ref{fig:discriminator_structure}.
The final sigmoid activation determines the probability that the input image is real or fake.
Note that our discriminator network does not employ the batch normalization \cite{ioffe2015batch}, because the batch size is too small to use that optimization.
In addition, it contains two more convolutional layers than the original SRGAN model due to the different size of the input image patches.

\subsection{Qualitative score predictors}
\label{sec:qualitative_score_predictors}

One of our main ideas for perceptually improved super-resolution is to utilize deep learning models classifying perceptual quality of images, instead of general image classifiers.
For this, we employ two deep networks that predict aesthetic and subjective quality scores of images, respectively.
To build the networks, we utilize the neural image assessment (NIMA) approach \cite{talebi2018nima}, which predicts the quality score of a given image.
This approach replaces the last layer of a well-known image classifier such as VGG \cite{simonyan2014very} or Inception-v3 \cite{szegedy2016rethinking} with a fully-connected layer with the softmax activation, which produces probabilities of 10 score classes.
In our implementation, MobileNetV2 \cite{sandler2018mobilenetv2} is used as the base image classifier, because it is much faster than the other image classifiers and supports various sizes of input images.

We build two score predictors: one for predicting \textit{aesthetic} scores and the other for predicting \textit{subjective} scores.
For the aesthetic score predictor, we employ the AVA dataset \cite{murray2012ava}, which contains aesthetic user ratings of the images shared in DPChallenge\footnote{\url{http://www.dpchallenge.com}}.
For the subjective score predictor, we use the TID2013 dataset \cite{ponomarenko2015image}, which consists of the subjective quality evaluation results for the test images degraded by various distortion types (e.g., compression, noise, and blurring).
While the AVA dataset provides exact score distributions, the TID2013 dataset only provides the mean and standard deviation of the scores.
Therefore, we approximate a Gaussian distribution with the mean and standard deviation to train the network based on TID2013.
In addition, we adjust the score range of the TID2013 dataset from $[0, 9]$ to $[1, 10]$ to match the range of the AVA dataset (i.e., $[1, 10]$).
After training the predictors, we use only the mean values of the predicted score distributions to enhance the perceptual quality of the upscaled images.

\subsection{Discussion}
\label{sec:structure_discussion}

The proposed model extends two existing networks: EUSR \cite{kim2018deep} as an upscaling model and SRGAN \cite{ledig2017photo} as a discriminator.
However, the two networks aim at different objectives: EUSR is for better quantitative quality and SRGAN is for better perceptual quality.
Our proposed model combines them to ensure both quantitative and perceptual quality, with two newly proposed components: multi-pass upscaling and qualitative score predictors.
In summary, our 4PP-EUSR model achieves the following benefits with the aforementioned components:
\begin{itemize}
	\item
	Our model can upscale the input images with considering both the quantitative and perceptual quality.
	While the base EUSR model tries to make the upscaled images similar to the ground-truth ones, the discriminator reinforces it to focus on fine details.
	Therefore, our model can achieve better quantitative quality than the other methods concentrating on perceptual quality while keeping the perceptual quality similar to theirs.
	We will thoroughly investigate this in Section~\ref{sec:existing_model_comparison}.\\
	
	\item
	Thanks to the multi-pass upscaling, the proposed model can learn various upscaling patterns, which will be further discussed in Sections~\ref{sec:comparing_upscaling_paths} and \ref{sec:multipass_effectiveness}.\\
	
	\item
	Employing the qualitative score predictors help our model generate perceptually improved images, since they are trained on the dataset that are obtained directly from human raters.
	We will discuss their benefits in Sections~\ref{sec:roles_of_loss_functions} and \ref{sec:comparing_different_loss_weights}.
\end{itemize}

\section{Training details}
\label{sec:training_details}

We train our model in three phases: pre-training the EUSR model, building qualitative score predictors, and training the EUSR model in a perceptual manner.
Our method is implemented on the TensorFlow framework \cite{abadi2016tensorflow}.

\subsection{Pre-training multi-scale super-resolution model}
\label{sec:pretraining_eusr}

In our method, we employ 32 and one local residual blocks in the residual module and the upscaling part of the EUSR model, respectively.
The EUSR model is first pre-trained with the training set of the DIV2K dataset \cite{timofte2018ntire} (i.e., 800 images) using the L1 reconstruction loss as in \cite{kim2018deep}.
For each training step, 16 image patches having a size of 48$\times$48 pixels are obtained by randomly cropping the training images.
Then, one of the upscaling paths (i.e., $\times$2, $\times$4, and $\times$8) is randomly selected and trained at that step.
For instance, when the $\times$2 path is selected, the parameters of the path of the model are trained to generate the upscaled images having a size of 96$\times$96 pixels.
The Adam optimization method \cite{kingma2014adam} with $\beta_{1}=0.9$, $\beta_{2}=0.999$, and $\hat{\epsilon}={10}^{-8}$ is used to update the parameters.
A total of 1,000,000 training steps are executed with an initial learning rate of ${10}^{-4}$ and reducing the learning rate by a half for every 200,000 steps.

\subsection{Training qualitative score predictors}

Along with pre-training EUSR, we also train the qualitative score predictors explained in Section~\ref{sec:qualitative_score_predictors}.
As the base image classifier, we employ MobileNetV2 \cite{sandler2018mobilenetv2} pre-trained on the ImageNet dataset \cite{russakovsky2015imagenet} with a width multiplier of 1.
In the original procedure of training NIMA \cite{talebi2018nima}, the input image is rescaled to 256$\times$256 pixels without considering the aspect ratio and then randomly cropped to 224$\times$224 pixels, which is the input size of VGG19 \cite{simonyan2014very} and Inception-v3 \cite{szegedy2016rethinking}.
However, these rescaling and cropping processes are not considered in our case because the MobileNetV2 model does not limit the size of an input image.
Instead, we set the input resolution of MobileNetV2 to 192$\times$192 pixels, which is the output size of the 4PP-EUSR model for input patches having a size of 48$\times$48 pixels.
In addition, we do not employ the rescaling step and only employ the cropping step to make the input image have a size of 192$\times$192 pixels, because the objective of our score predictors is to evaluate the quality of patches, not the whole given image.

As the loss function for training the qualitative score predictors, we employ the squared Earth mover's distance defined in \cite{hou2016squared} as
\begin{equation}
\label{eq:squared_earth_movers_distance}
{E}( {Q}^{I}, {Q}^{\widetilde{I}} ) = \sum_{i} \Big( {F}_{i}({Q}^{I}) - {F}_{i}({Q}^{\widetilde{I}}) \Big) ^{2}
\end{equation}
where $I$ and $\widetilde{I}$ are the ground-truth and upscaled images, respectively, ${Q}^{I}$ and ${Q}^{\widetilde{I}}$ are the probability distributions of the qualitative scores obtained from the predictor for the two images, respectively, and ${F}_{i}( \cdot )$ is the $i$-th element of the cumulative distribution function of the input.
The Adam optimization method \cite{kingma2014adam} with $\beta_{1}=0.9$, $\beta_{2}=0.999$, and $\epsilon={10}^{-7}$ is used to train the parameters.

For the aesthetic score predictor, we use about 5,000 images of the AVA dataset \cite{murray2012ava} for validation and the remaining 250,000 images for training.
We first train the new last fully-connected layer for five epochs with a batch size of 128 and a learning rate of ${10}^{-3}$, while freezing all other layers.
Then, all the layers are fine-tuned for five epochs with a batch size of 32 and a learning rate of ${10}^{-5}$.
For the validation images cropped in the center parts, the predictor achieves an average squared Earth mover's distance of 0.079. 

For the subjective score predictor, we consider the first three reference images and their degraded versions in the TID2013 dataset \cite{ponomarenko2015image} (corresponding to 360 score distributions) for validation and the remaining 22 reference images and their degraded versions (corresponding to 2,640 score distributions) for training.
Similarly to the aesthetic score predictor, we first train the subjective score predictor with freezing all the layers except the new last fully-connected layer for 100 epochs with a batch size of 128 and a learning rate of ${10}^{-3}$.
Then, the whole network is fine-tuned for 100 epochs with a batch size of 32 and a learning rate of ${10}^{-5}$.
For the validation images cropped in the center parts, the predictor achieves a Spearman's rank correlation coefficient (SROCC) of 0.780.

\subsection{Training super-resolution model}
\label{sec:training_perceptual_sr}

Finally, we fine-tune the pre-trained EUSR model together with the discriminator network using the two trained qualitative score predictors.
At each training step, the 4PP-EUSR model outputs three upscaled images by a factor of 4.
Then, the discriminator is trained to differentiate the ground-truth and upscaled images based on the sigmoid cross entropy loss as in \cite{ledig2017photo}.
After updating parameters of the discriminator, the 4PP-EUSR model is trained with six losses defined as follows.
\begin{itemize}
	\item \textbf{Reconstruction loss (${l}_{r}$).}
	The reconstruction loss represents the main objective of the super-resolution task: each pixel value of the super-resolved image must be as close as possible to that of the ground-truth image.
	In our model, this loss is measured by the pixel-by-pixel L1 loss between the ground-truth and generated images, i.e.,
	\begin{equation}
	{l}_{r} = \frac{1}{W \times H} \sum_{w} \sum_{h} \left| {I}_{w, h} - \widetilde{{I}}_{w, h} \right|
	\end{equation}
	where $W$ and $H$ are the width and height of the images, respectively, and ${I}_{w, h}$ and $\widetilde{{I}}_{w, h}$ are the pixel values at $(w, h)$ of the ground-truth and upscaled images, respectively.\\
	
	\item \textbf{Adversarial loss (${l}_{g}$).}
	The output of the discriminator network is used to train the super-resolution model towards enhancing perceptual quality, which is denoted as the adversarial loss.
	It is calculated by the sigmoid cross entropy of the logits obtained from the discriminator for the upscaled images \cite{ledig2017photo}:
	\begin{equation}
	{l}_{g} = - \log({D}^{\widetilde{I}})
	\end{equation}
	where ${D}^{\widetilde{I}}$ is the output of the discriminator for the upscaled image $\widetilde{I}$, which represents the probability that the given image is a real one.\\
	
	\item \textbf{Aesthetic score loss (${l}_{as}$).}
	We obtain the aesthetic scores of both the ground-truth and upscaled images from the trained aesthetic score predictor.
	Then, we define the aesthetic score loss as the weighted difference between the scores, i.e.,
	\begin{equation}
	{l}_{as} = \max \big( 0, ({S}_{a,\textrm{max}} - {S}_{a}^{\widetilde{I}}) - \alpha_{as} ({S}_{a,\textrm{max}} - {S}_{a}^{I}) \big)
	\end{equation}
	where ${S}_{a}^{I}$ and ${S}_{a}^{\widetilde{I}}$ are the predicted aesthetic scores of the ground-truth and upscaled images, respectively.
	${S}_{a,\textrm{max}}$ is the maximum aesthetic score, which is 10 in our case.
	The term $\alpha_{as}$ plays a role to control the expected level of aesthetic quality of the upscaled image.
	For example, $\alpha_{as} < 1.0$ enforces the model to generate an image that is even perceptually better than the ground-truth image.
	In our experiments, we set $\alpha_{as}$ to 0.8.\\
	
	\item \textbf{Aesthetic representation loss (${l}_{ar}$).}
	Inspired by \cite{ledig2017photo}, we also define the aesthetic representation loss, which is the L2 loss between the intermediate outputs of the ``global average pooling'' layer in the aesthetic score predictor for both the ground-truth and upscaled images:
	\begin{equation}
	{l}_{ar} = \sum_{i} {\big({P}_{a, i}^{I} - {P}_{a, i}^{\widetilde{I}}\big)}^{2}
	\end{equation}
	where ${P}_{a, i}^{I}$ and ${P}_{a, i}^{\widetilde{I}}$ are the $i$-th values of the intermediate outputs for the ground-truth and upscaled images, respectively.
	The length of each intermediate output is 1,280 \cite{sandler2018mobilenetv2}.\\
	
	\item \textbf{Subjective score loss (${l}_{ss}$).}
	In the same manner as the aesthetic score loss, we calculate the subjective score loss using the trained subjective score predictor, i.e.,
	\begin{equation}
	{l}_{ss} = \max \big( 0, ({S}_{s,\textrm{max}} - {S}_{s}^{\widetilde{I}}) - \alpha_{ss} ({S}_{s,\textrm{max}} - {S}_{s}^{I}) \big)
	\end{equation}
	where ${S}_{s}^{I}$ and ${S}_{s}^{\widetilde{I}}$ are the predicted subjective scores of the ground-truth and upscaled images, respectively.
	${S}_{s,\textrm{max}}$ is the maximum subjective score, which is 10 in our case.
	Similarly to $\alpha_{as}$, the term $\alpha_{ss}$ controls the contribution of ${S}_{s}^{I}$, which is set to 0.8 in our experiments.\\
	
	\item \textbf{Subjective representation loss (${l}_{sr}$).}
	In the same manner as the aesthetic representation loss, we calculate the subjective representation loss using the subjective score predictor as
	\begin{equation}
	{l}_{sr} = \sum_{i} {\big({P}_{s, i}^{I} - {P}_{s, i}^{\widetilde{I}}\big)}^{2}
	\end{equation}
	where ${P}_{s, i}^{I}$ and ${P}_{s, i}^{\widetilde{I}}$ are the $i$-th values of the intermediate outputs at the ``global average pooling'' layer for the ground-truth and upscaled images, respectively.
	
\end{itemize}
The losses are calculated for all the three upscaled images and then averaged.

We use the 800 training images of the DIV2K dataset as in the pre-training phase.
The Adam optimization method \cite{kingma2014adam} with $\beta_{1}=0.9$, $\beta_{2}=0.999$, and $\hat{\epsilon}={10}^{-8}$ is used to train both the 4PP-EUSR and discriminator.
At every training step, two input image patches are selected, which results in generating six upscaled images.
Thus, the effective batch sizes of the upscaling and discriminative models are six and eight (i.e., two ground-truth and six upscaled images), respectively.
A total of $4 \times {10}^{5}$ steps are executed with learning rates of ${10}^{-5}$ and $2 \times {10}^{-5}$ for the 4PP-EUSR and discriminator, respectively.

\section{Results}
\label{sec:results}

In this section, we report the results of five experiments: comparing the performance of our method and other state-of-the-art super-resolution models, comparing the outputs obtained from different upscaling paths, comparing the performance of our method trained with and without multi-pass upscaling, investigating the roles of loss functions, and comparing the results obtained from different combinations of the loss weights.
For the first four experiments, we train our model with the following weighted sum of the six losses defined in Section~\ref{sec:training_perceptual_sr}:
\begin{equation}
\label{eq:default_loss_equation}
l = 0.05 {l}_{r} + 0.1 {l}_{g} + 0.01 {l}_{as} + 0.1 {l}_{ar} + 0.01 {l}_{ss} + 0.1 {l}_{sr}
\end{equation}
which is empirically determined to ensure high perceptual improvement with minimizing degradation of quantitative performance.

We evaluate the super-resolution performance on the Set5 \cite{bevilacqua2012low}, Set14 \cite{zeyde2010single}, and BSD100 \cite{martin2001database} datasets.
Each dataset contains 4, 14, and 100 images, respectively.
We employ five performance metrics that are widely used in the literature, including PSNR, SSIM \cite{wang2004image}, NIQE \cite{mittal2013making}, a no-reference super-resolution (SR) score proposed by Ma \textit{et al.} \cite{ma2017learning}, and perceptual index (PI) \cite{blau20182018}.
PSNR and SSIM are for measuring the quantitative quality, and higher values mean better quality.
NIQE, the SR score, and PI are for measuring the perceptual quality, and PI is obtained from the combination of NIQE and the SR score, i.e.,
\begin{equation}
\label{eq:perceptual_index}
\mathrm{PI}(\tilde{I}) = \frac{1}{2} \big( ( 10 - \mathrm{SR}(\tilde{I}) ) + \mathrm{NIQE}(\tilde{I}) \big)
\end{equation}
where $\tilde{I}$ is a given upscaled image, and $\mathrm{NIQE}(\cdot)$ and $\mathrm{SR}(\cdot)$ are the measured NIQE value and the SR score, respectively.
For NIQE and PI, lower values mean better quality.
For the SR score, higher values mean better quality.
All quality metrics are calculated on the Y channel of the YCbCr channels converted from the RGB channels with cropping 4 pixels of each border, as in many existing studies \cite{ledig2017photo,kim2018deep,lim2017enhanced}.
In addition, we conduct a subjective test to assess the performance of our method in the perspective of real human observers.

\subsection{Comparison with existing models}
\label{sec:existing_model_comparison}

We first compare the result images obtained from the $\times$4 path of our model with those by the following existing super-resolution models.
\begin{itemize}
	\item \textbf{Bicubic interpolation.}
	It is a traditional upscaling method, which interpolates pixel values based on values of their adjacent pixels.\\
	
	\item \textbf{SRResNet \cite{ledig2017photo}.}
	This is for single-scale super-resolution, which consists of several residual blocks.
	Its two variants are considered: The SRResNet-MSE model is trained with the mean-squared loss and the SRResNet-VGG22 model is trained with the Euclidean distance-based loss for the output of the second \textit{conv3-128} layer of VGG19.
	Their results are retrieved from the authors' supplementary material\footnote{\url{https://twitter.box.com/s/lcue6vlrd01ljkdtdkhmfvk7vtjhetog}}.\\
	
	\item \textbf{EDSR \cite{lim2017enhanced}.}
	This model also consists of residual blocks similarly to SRResNet, but does not employ batch normalization to improve the performance.
	In addition, the upscaled results are obtained by a so-called ``geometric self-ensemble'' strategy, which obtains eight geometrically transformed versions of the input image via flipping and rotation and blends the model outputs for them.
	The compared results are obtained from a model trained on the DIV2K dataset, which is provided by the authors\footnote{\url{https://cv.snu.ac.kr/research/EDSR/model_pytorch.tar}}.\\
	
	\item \textbf{MDSR \cite{lim2017enhanced}.}
	It is an extended version of EDSR, which supports multiple factors of upscaling.
	We obtain the upscaled images from the $\times$4 path of the MDSR model trained on the DIV2K dataset \cite{timofte2018ntire}.
	The trained model is provided by the authors\footnote{\url{https://cv.snu.ac.kr/research/EDSR/model_pytorch.tar}}.\\
	
	\item \textbf{EUSR \cite{kim2018deep}.}
	This is the base model of 4PP-EUSR, which supports multi-scale super-resolution and consists of optimized residual modules as explained in Section~\ref{sec:eusr}.
	We consider the pre-trained EUSR model described in Section~\ref{sec:pretraining_eusr} as a baseline.\\
	
	\item \textbf{RCAN \cite{zhang2018image}.}
	The RCAN model employs a channel attention mechanism along with the residual-in-residual structure, which contributes to output better super-resolved images than EDSR in terms of PSNR.
	In addition, RCAN employs the self-ensemble strategy to improve the performance as in the EDSR model.
	We obtain the output images from the trained model provided by the authors\footnote{\url{https://github.com/yulunzhang/RCAN}}.	\\
	
	\item \textbf{SRGAN \cite{ledig2017photo}.}
	The SRGAN model is an extended version of the SRResNet model, where a discriminator network is added to improve the perceptual quality of the upscaled outputs.
	We consider three SRGAN models, which use different loss functions to train the discriminator: SRGAN-MSE (the mean-squared loss), SRGAN-VGG22 (the Euclidean distance-based loss for the output of the second \textit{conv3-128} layer of VGG19), and SRGAN-VGG54 (the Euclidean distance-based loss for the output of the fourth \textit{conv3-512} layer of VGG19).
	The compared results are retrieved from the authors' supplementary material\footnote{\url{https://twitter.box.com/s/lcue6vlrd01ljkdtdkhmfvk7vtjhetog}}.\\
	
	\item \textbf{CX \cite{mechrez2018learning}.}
	This model is based on SRGAN but employs an additional loss function, the contextual loss \cite{mechrez2018contextual}, which measures the cosine distance between the VGG19 features for the ground-truth and upscaled images.
	The compared results are retrived from the authors' website\footnote{\url{http://cgm.technion.ac.il/people/Roey/index.html}}.
	
\end{itemize}

\begin{table*}[t]
	\scriptsize
	\centering
	\caption{Properties of the baseline and our models with respect to the number of parameters, multi-scale structure, and loss functions.}
	\label{table:baseline_properties}
	\begin{tabular}{l C{1.4cm} C{1.4cm} C{1.4cm} C{1.4cm} C{1.4cm} C{1.4cm}}
		Models & \# parameters & Multi-scale structure & Using reconstruction loss & Using adversarial loss & Using feature-based loss & Using perceptual loss \\
		\noalign{\smallskip}
		\hline
		\noalign{\smallskip}
		SRResNet-MSE & 1.5M & No & Yes & No & No & No \\
		SRResNet-VGG22 & 1.5M & No & No & No & Yes & No \\
		EDSR & 43.1M & No & Yes & No & No & No \\
		MDSR & 8.0M & Yes & Yes & No & No & No \\
		EUSR & 6.3M & Yes & Yes & No & No & No \\
		RCAN & 15.6M & No & Yes & No & No & No \\
		SRGAN-MSE & 1.5M & No & Yes & Yes & No & No \\
		SRGAN-VGG22 & 1.5M & No & Yes\textsuperscript{\dag} & Yes & Yes & No \\
		SRGAN-VGG54 & 1.5M & No & Yes\textsuperscript{\dag} & Yes & Yes & No \\
		CX & 1.5M & No & Yes & Yes & Yes & No \\
		\textbf{4PP-EUSR (Ours)} & 6.3M & Yes & Yes & Yes & Yes & Yes
	\end{tabular}
	\begin{tablenotes}
		\item \textsuperscript{\dag}~For pre-training
	\end{tablenotes}
\end{table*}

Table~\ref{table:baseline_properties} compares properties of the baselines and ours, including the number of model parameters, the existence of a multi-scale structure, whether to use the reconstruction loss, whether to employ the discriminator, whether to compare features obtained from well-known image classifiers (e.g., VGG19), and whether to use perceptual scores.
First, the EDSR model consists of the largest number of parameters than the other models, while the SRResNet, SRGAN, and CX models have the smallest number of parameters.
Our model contains a smaller number of parameters than the MDSR and RCAN models.
In terms of the multi-scale structure, MDSR, EUSR, and our model utilize multiple scaling factors, while the other models are based on single-scale super-resolution.
Although all the models except SRResNet-VGG22 employ the reconstruction loss, the SRGAN-VGG22 and SRGAN-VGG54 models use it only for pre-training.
In addition, SRGANs, CX, and our model employ discriminator networks and use them for adversarial losses.
SRResNet-MSE, SRGAN-VGG22, SRGAN-VGG54, and CX employ VGG19 as an additional network to use its intermediate outputs as feature-based losses.
Our model employs the MobileNetV2-based networks instead of VGG19.
Finally, ours estimates the aesthetic and subjective quality scores of the ground-truth and upscaled images for calculating perceptual losses.

\begin{table}[t]
	\setlength{\tabcolsep}{0.4em}
	\scriptsize
	\centering
	\caption{Performance comparison of the baselines and our model evaluated on the Set5 \cite{bevilacqua2012low}, Set14 \cite{zeyde2010single}, and BSD100 \cite{martin2001database} datasets. The models are sorted by PSNR (dB) in an ascending order.}
	\label{table:result_baseline_comparison}
	\begin{tabular}{lccccc}
		\textbf{Set5} & PSNR (dB) & SSIM & NIQE & SR score & PI \\
		\noalign{\smallskip}
		\hline
		\noalign{\smallskip}
		Bicubic & 28.418 & 0.810 & 8.540 & 3.770 & 7.385 \\
		CX & 29.102 & 0.830 & 4.546 & 7.957 & 3.295 \\
		SRGAN-VGG54 & 29.410 & 0.834 & 4.651 & 7.940 & 3.355 \\
		SRGAN-VGG22 & 29.871 & 0.835 & 4.919 & 7.534 & 3.692 \\
		SRResNet-VGG22 & 30.501 & 0.869 & 6.905 & 6.336 & 5.285 \\
		SRGAN-MSE & 30.666 & 0.859 & 4.997 & 7.308 & 3.844 \\
		\textbf{4PP-EUSR (Ours)} & 31.369 & 0.870 & 5.366 & 6.890 & 4.238 \\
		SRResNet-MSE & 32.058 & 0.892 & 7.194 & 5.411 & 5.891 \\
		EUSR & 32.352 & 0.896 & 7.070 & 5.173 & 5.949 \\
		MDSR & 32.533 & 0.898 & 7.111 & 5.109 & 6.001 \\
		EDSR & 32.630 & 0.899 & 7.235 & 5.211 & 6.012 \\
		RCAN & 32.713 & 0.899 & 7.229 & 5.277 & 5.976 \\
		\noalign{\smallskip}
		\noalign{\smallskip}
		\textbf{Set14} & PSNR (dB) & SSIM & NIQE & SR score & PI \\
		\noalign{\smallskip}
		\hline
		\noalign{\smallskip}
		CX & 26.011 & 0.700 & 3.460 & 7.942 & 2.759 \\
		Bicubic & 26.091 & 0.705 & 7.764 & 3.661 & 7.051 \\
		SRGAN-VGG54 & 26.114 & 0.696 & 3.875 & 8.111 & 2.882 \\
		SRGAN-VGG22 & 26.529 & 0.712 & 4.221 & 7.983 & 3.119 \\
		SRGAN-MSE & 27.006 & 0.719 & 4.005 & 7.877 & 3.064 \\
		SRResNet-VGG22 & 27.272 & 0.742 & 7.023 & 7.093 & 4.965 \\
		\textbf{4PP-EUSR (Ours)} & 27.969 & 0.751 & 4.147 & 7.457 & 3.345 \\
		SRResNet-MSE & 28.590 & 0.782 & 6.075 & 5.648 & 5.213 \\
		EUSR & 28.750 & 0.786 & 6.168 & 5.467 & 5.351 \\
		MDSR & 28.895 & 0.789 & 6.267 & 5.311 & 5.478 \\
		RCAN & 28.951 & 0.790 & 6.343 & 5.451 & 5.446 \\
		EDSR & 28.953 & 0.790 & 6.305 & 5.379 & 5.463 \\
		\noalign{\smallskip}
		\noalign{\smallskip}
		\textbf{BSD100} & PSNR (dB) & SSIM & NIQE & SR score & PI \\
		\noalign{\smallskip}
		\hline
		\noalign{\smallskip}
		CX & 24.581 & 0.644 & 3.301 & 8.801 & 2.250 \\
		SRGAN-VGG54 & 25.176 & 0.641 & 3.407 & 8.705 & 2.351 \\
		SRGAN-VGG22 & 25.697 & 0.660 & 3.750 & 8.488 & 2.631 \\
		Bicubic & 25.957 & 0.669 & 7.712 & 3.723 & 6.995 \\
		SRGAN-MSE & 25.981 & 0.643 & 4.032 & 8.428 & 2.802 \\
		SRResNet-VGG22 & 26.322 & 0.694 & 7.805 & 7.439 & 5.183 \\
		\textbf{4PP-EUSR (Ours)} & 26.904 & 0.701 & 3.820 & 7.907 & 2.956 \\
		SRResNet-MSE & 27.601 & 0.737 & 6.240 & 5.807 & 5.217 \\
		EUSR & 27.674 & 0.740 & 6.423 & 5.808 & 5.307 \\
		MDSR & 27.771 & 0.743 & 6.538 & 5.690 & 5.424 \\
		EDSR & 27.796 & 0.744 & 6.432 & 5.779 & 5.326 \\
		RCAN & 27.821 & 0.745 & 6.451 & 5.868 & 5.292
	\end{tabular}
\end{table}

\begin{figure}[t]
	\centering
	\includegraphics[width=0.6\linewidth]{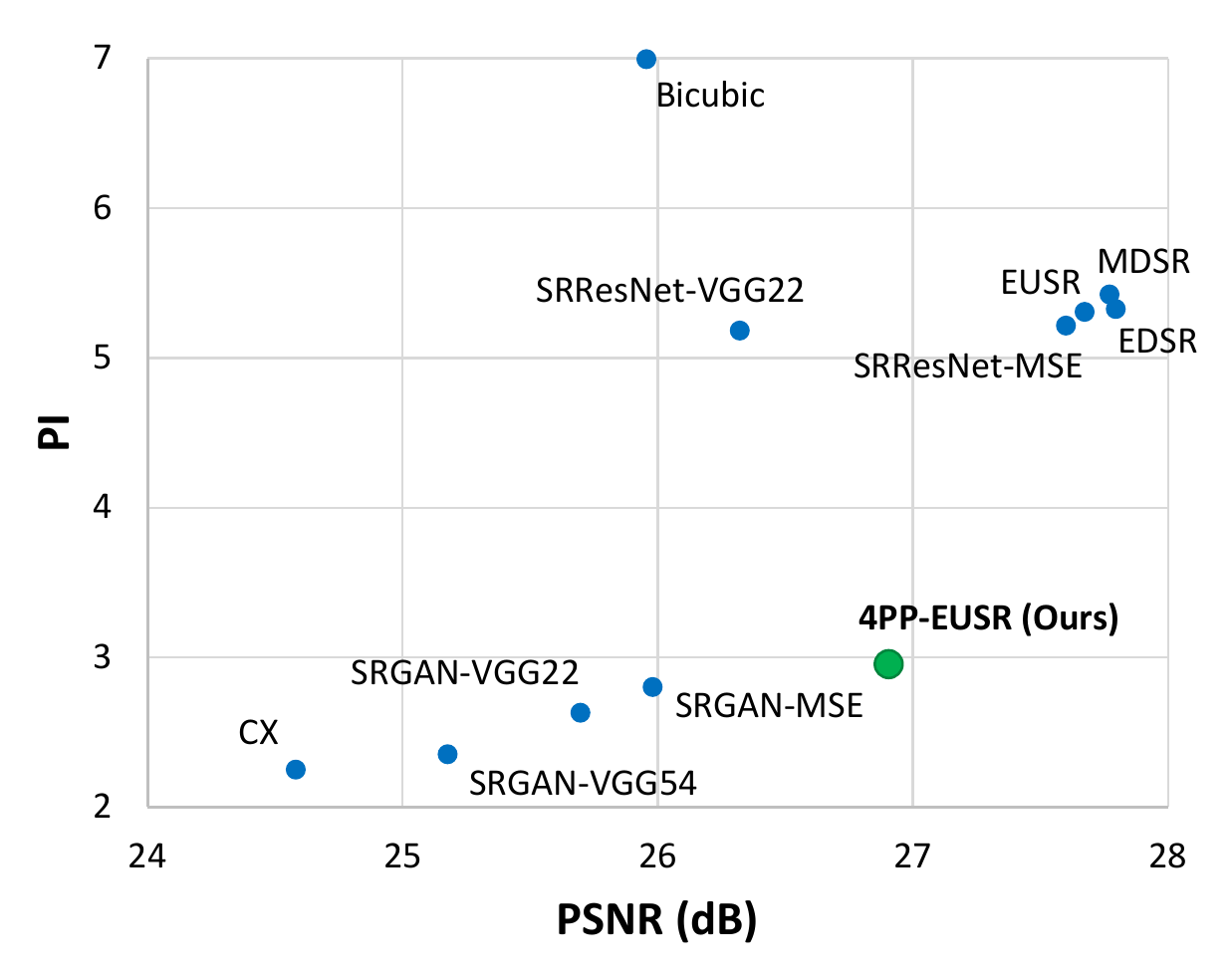}
	\caption{PSNR and PI values of the baselines and our model for the BSD100 dataset \cite{martin2001database}.}
	\label{fig:psnr_pi_bsd100}
\end{figure}

Table~\ref{table:result_baseline_comparison} shows the performance comparison of the baselines and ours evaluated on the three datasets.
First of all, the bicubic interpolation introduces a large amount of distortion, which results in low PSNR values, and the upscaled images have poor perceptual quality, according to the high PI values.
The models that do not employ a discriminator network (i.e., SRResNet, EDSR, MDSR, EUSR, and RCAN) achieve better quantitative quality than the others, showing higher PSNR values, but their perceptual quality is worse except the bicubic interpolation, showing higher PI values.
The models considering perceptual quality (i.e., SRGAN and CX) have similar or only slightly higher PSNR values in comparison to the bicubic interpolation, but their perceptual quality is far better than that of the bicubic interpolation, according to the much lower PI values.
Our model (i.e., 4PP-EUSR) always records PSNR values higher than those of the other discriminator-based models, which means that ours generates quantitatively better upscaled outputs.
At the same time, our model achieves perceptual quality similar to that of SRGAN-MSE in terms of the PI value.
For instance, for the BSD100 dataset, the PI values of our model and SRGAN-MSE are 2.956 and 2.802, respectively.
This appears more clearly in \figurename~\ref{fig:psnr_pi_bsd100}, which compares the baselines and our model with respect to PSNR and PI values measured for the BSD100 dataset.
It confirms that our model achieves proper balances of the quantitative and qualitative quality of the upscaled images.

\begin{figure*}[]
	\centering
	\begin{minipage}[b]{0.24\linewidth}
		\centering
		\centerline{\scriptsize{Bicubic}}\medskip
		\centerline{\includegraphics[width=0.98\linewidth]{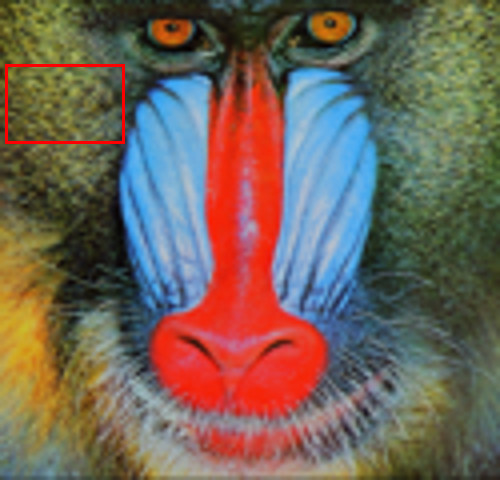}}\smallskip
		\centerline{\includegraphics[width=0.98\linewidth]{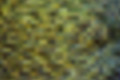}}
	\end{minipage}
	\begin{minipage}[b]{0.24\linewidth}
		\centering
		\centerline{\scriptsize{SRResNet-MSE}}\medskip
		\centerline{\includegraphics[width=0.98\linewidth]{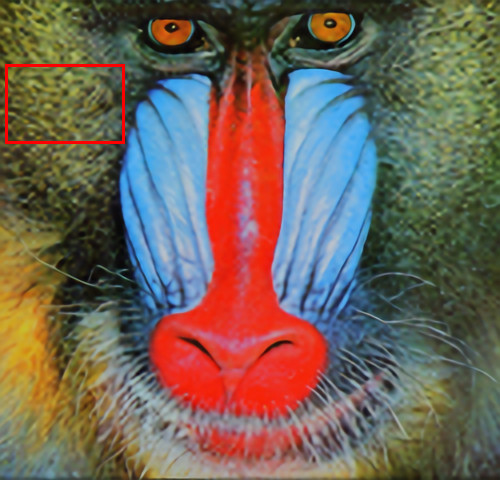}}\smallskip
		\centerline{\includegraphics[width=0.98\linewidth]{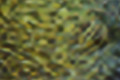}}
	\end{minipage}
	\begin{minipage}[b]{0.24\linewidth}
		\centering
		\centerline{\scriptsize{SRResNet-VGG22}}\medskip
		\centerline{\includegraphics[width=0.98\linewidth]{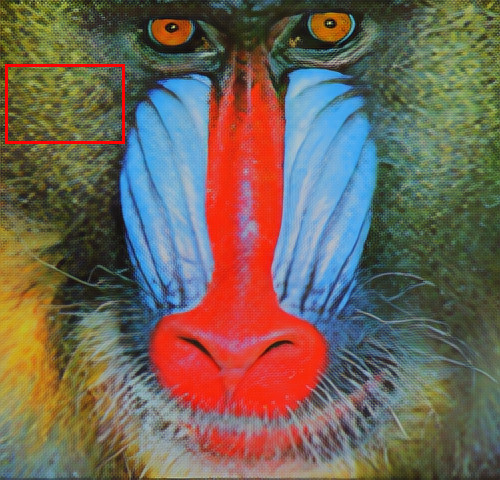}}\smallskip
		\centerline{\includegraphics[width=0.98\linewidth]{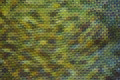}}
	\end{minipage}
	\begin{minipage}[b]{0.24\linewidth}
		\centering
		\centerline{\scriptsize{EDSR}}\medskip
		\centerline{\includegraphics[width=0.98\linewidth]{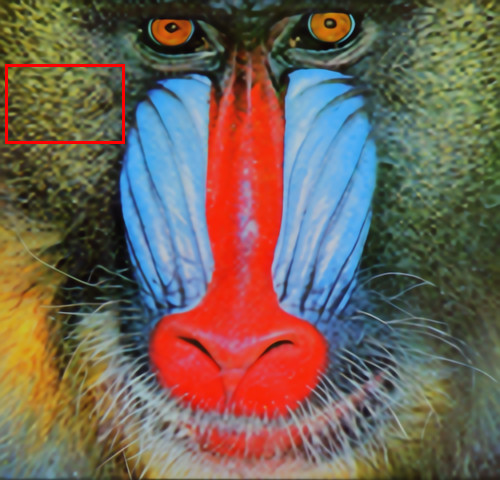}}\smallskip
		\centerline{\includegraphics[width=0.98\linewidth]{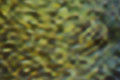}}
	\end{minipage}
	\medskip \\ \medskip
	\begin{minipage}[b]{0.24\linewidth}
		\centering
		\centerline{\scriptsize{MDSR}}\medskip
		\centerline{\includegraphics[width=0.98\linewidth]{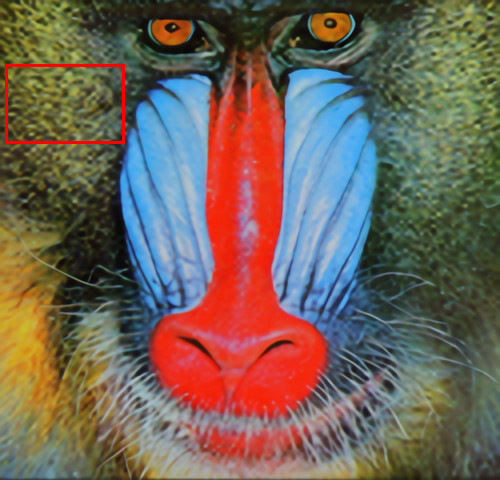}}\smallskip
		\centerline{\includegraphics[width=0.98\linewidth]{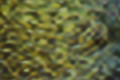}}
	\end{minipage}
	\begin{minipage}[b]{0.24\linewidth}
		\centering
		\centerline{\scriptsize{EUSR}}\medskip
		\centerline{\includegraphics[width=0.98\linewidth]{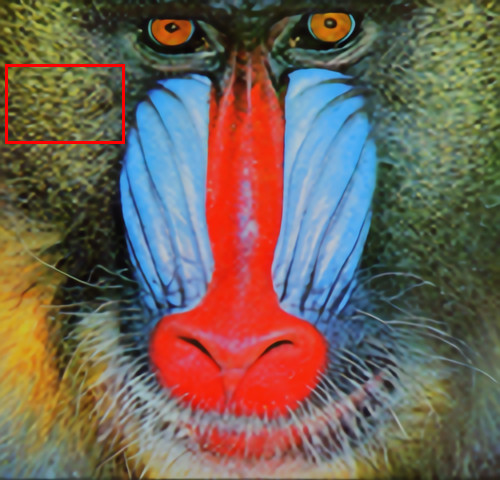}}\smallskip
		\centerline{\includegraphics[width=0.98\linewidth]{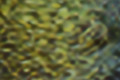}}
	\end{minipage}
	\begin{minipage}[b]{0.24\linewidth}
		\centering
		\centerline{\scriptsize{RCAN}}\medskip
		\centerline{\includegraphics[width=0.98\linewidth]{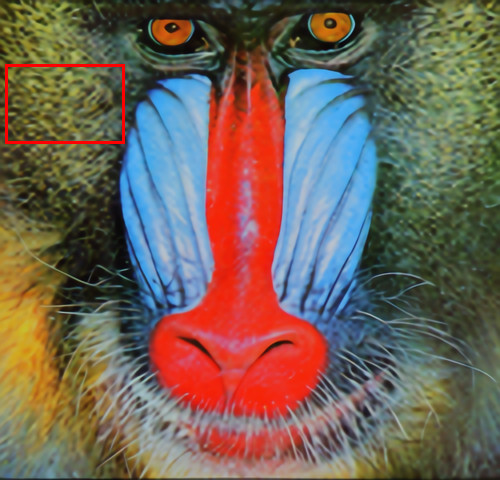}}\smallskip
		\centerline{\includegraphics[width=0.98\linewidth]{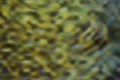}}
	\end{minipage}
	\begin{minipage}[b]{0.24\linewidth}
		\centering
		\centerline{\scriptsize{SRGAN-MSE}}\medskip
		\centerline{\includegraphics[width=0.98\linewidth]{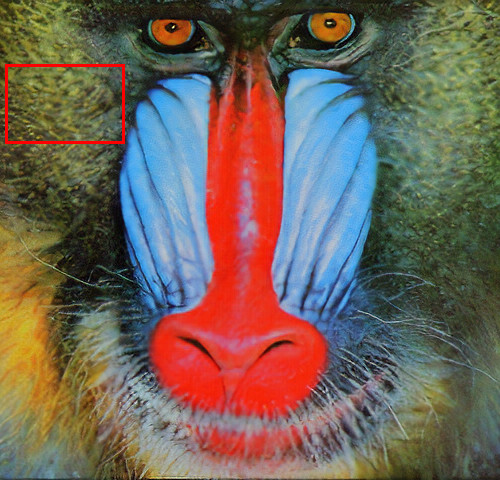}}\smallskip
		\centerline{\includegraphics[width=0.98\linewidth]{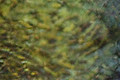}}
	\end{minipage}
	\medskip \\ \medskip
	\begin{minipage}[b]{0.24\linewidth}
		\centering
		\centerline{\scriptsize{SRGAN-VGG22}}\medskip
		\centerline{\includegraphics[width=0.98\linewidth]{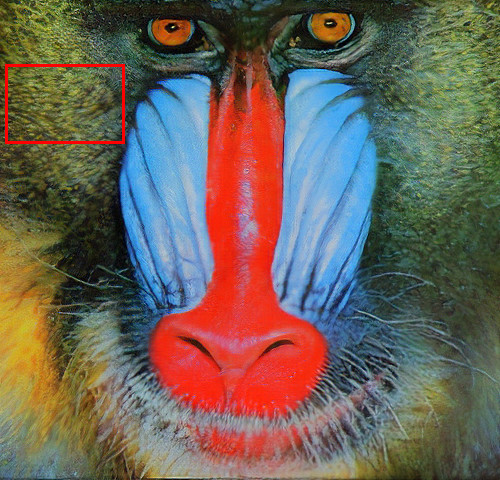}}\smallskip
		\centerline{\includegraphics[width=0.98\linewidth]{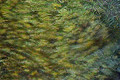}}
	\end{minipage}
	\begin{minipage}[b]{0.24\linewidth}
		\centering
		\centerline{\scriptsize{SRGAN-VGG54}}\medskip
		\centerline{\includegraphics[width=0.98\linewidth]{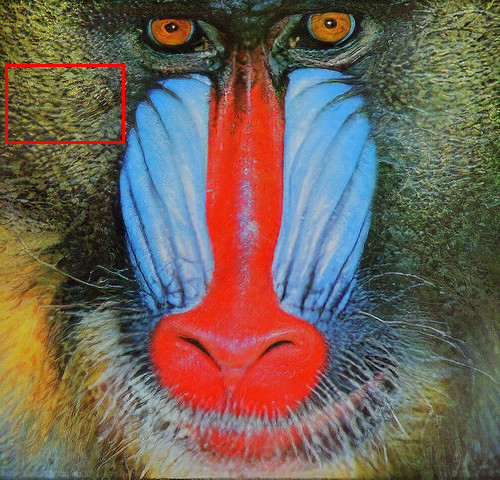}}\smallskip
		\centerline{\includegraphics[width=0.98\linewidth]{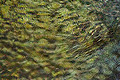}}
	\end{minipage}
	\begin{minipage}[b]{0.24\linewidth}
		\centering
		\centerline{\scriptsize{CX}}\medskip
		\centerline{\includegraphics[width=0.98\linewidth]{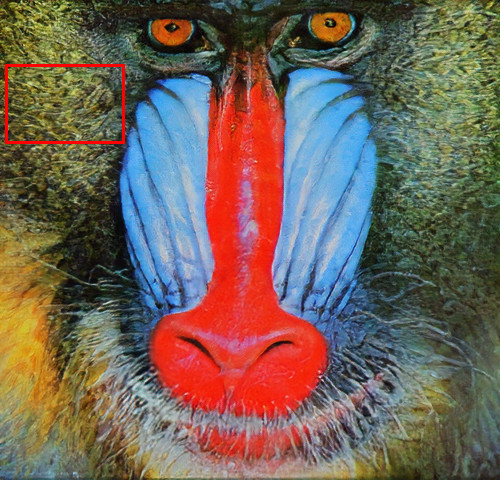}}\smallskip
		\centerline{\includegraphics[width=0.98\linewidth]{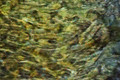}}
	\end{minipage}
	\begin{minipage}[b]{0.24\linewidth}
		\centering
		\centerline{\scriptsize{4PP-EUSR (Ours)}}\medskip
		\centerline{\includegraphics[width=0.98\linewidth]{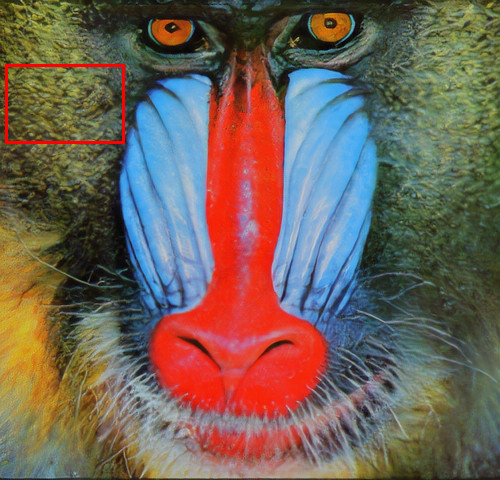}}\smallskip
		\centerline{\includegraphics[width=0.98\linewidth]{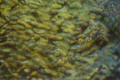}}
	\end{minipage}
	\caption{Images reconstructed by the baselines and our model. The input images are from the Set14 dataset \cite{zeyde2010single}.}
	\label{fig:result_baseline_comparison}
\end{figure*}

\begin{figure*}[]
	\centering
	\begin{minipage}[b]{0.24\linewidth}
		\centering
		\centerline{\scriptsize{Bicubic}}\medskip
		\centerline{\includegraphics[width=0.98\linewidth]{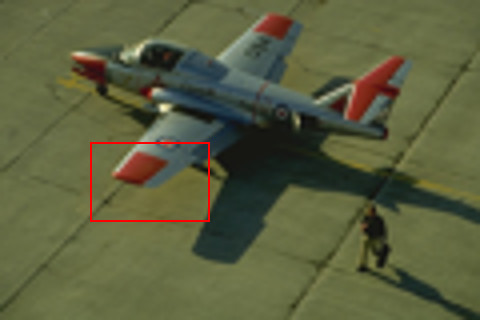}}\smallskip
		\centerline{\includegraphics[width=0.98\linewidth]{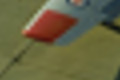}}
	\end{minipage}
	\begin{minipage}[b]{0.24\linewidth}
		\centering
		\centerline{\scriptsize{SRResNet-MSE}}\medskip
		\centerline{\includegraphics[width=0.98\linewidth]{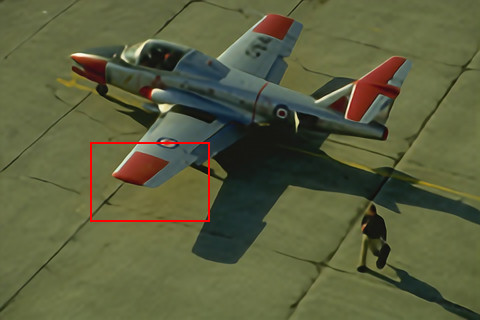}}\smallskip
		\centerline{\includegraphics[width=0.98\linewidth]{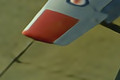}}
	\end{minipage}
	\begin{minipage}[b]{0.24\linewidth}
		\centering
		\centerline{\scriptsize{SRResNet-VGG22}}\medskip
		\centerline{\includegraphics[width=0.98\linewidth]{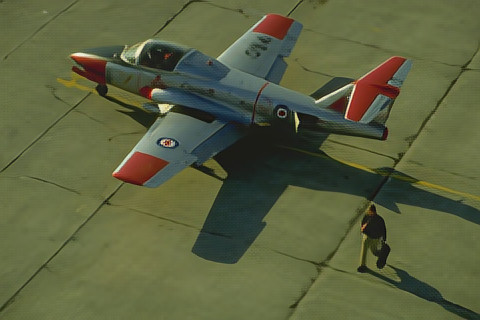}}\smallskip
		\centerline{\includegraphics[width=0.98\linewidth]{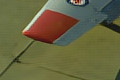}}
	\end{minipage}
	\begin{minipage}[b]{0.24\linewidth}
		\centering
		\centerline{\scriptsize{EDSR}}\medskip
		\centerline{\includegraphics[width=0.98\linewidth]{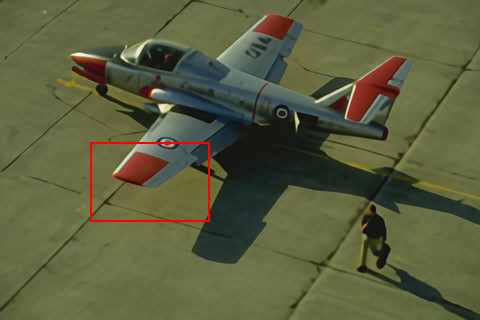}}\smallskip
		\centerline{\includegraphics[width=0.98\linewidth]{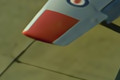}}
	\end{minipage}
	\medskip \\ \medskip
	\begin{minipage}[b]{0.24\linewidth}
		\centering
		\centerline{\scriptsize{MDSR}}\medskip
		\centerline{\includegraphics[width=0.98\linewidth]{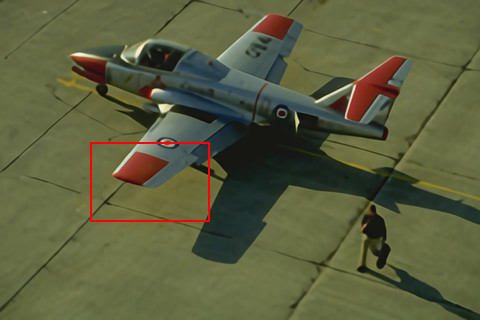}}\smallskip
		\centerline{\includegraphics[width=0.98\linewidth]{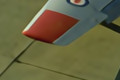}}
	\end{minipage}
	\begin{minipage}[b]{0.24\linewidth}
		\centering
		\centerline{\scriptsize{EUSR}}\medskip
		\centerline{\includegraphics[width=0.98\linewidth]{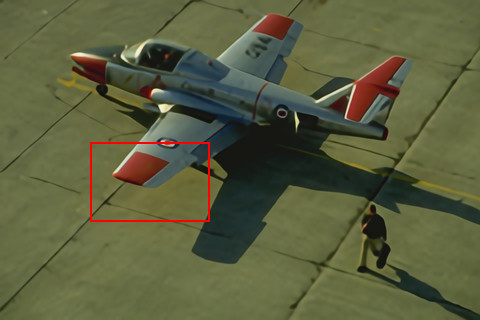}}\smallskip
		\centerline{\includegraphics[width=0.98\linewidth]{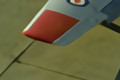}}
	\end{minipage}
	\begin{minipage}[b]{0.24\linewidth}
		\centering
		\centerline{\scriptsize{RCAN}}\medskip
		\centerline{\includegraphics[width=0.98\linewidth]{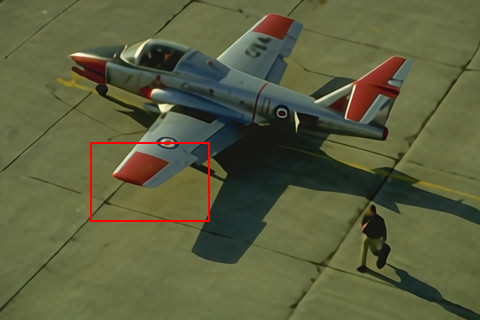}}\smallskip
		\centerline{\includegraphics[width=0.98\linewidth]{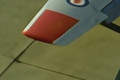}}
	\end{minipage}
	\begin{minipage}[b]{0.24\linewidth}
		\centering
		\centerline{\scriptsize{SRGAN-MSE}}\medskip
		\centerline{\includegraphics[width=0.98\linewidth]{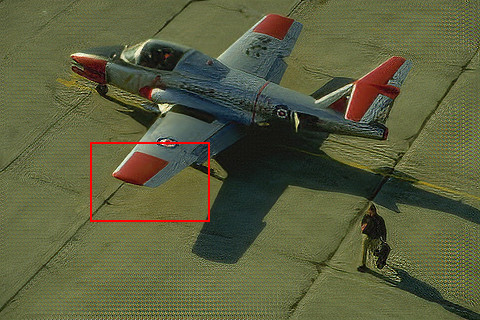}}\smallskip
		\centerline{\includegraphics[width=0.98\linewidth]{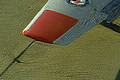}}
	\end{minipage}
	\medskip \\ \medskip
	\begin{minipage}[b]{0.24\linewidth}
		\centering
		\centerline{\scriptsize{SRGAN-VGG22}}\medskip
		\centerline{\includegraphics[width=0.98\linewidth]{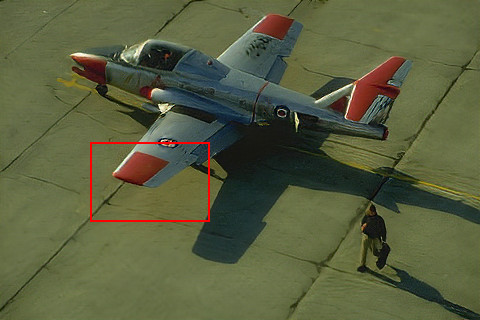}}\smallskip
		\centerline{\includegraphics[width=0.98\linewidth]{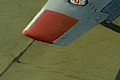}}
	\end{minipage}
	\begin{minipage}[b]{0.24\linewidth}
		\centering
		\centerline{\scriptsize{SRGAN-VGG54}}\medskip
		\centerline{\includegraphics[width=0.98\linewidth]{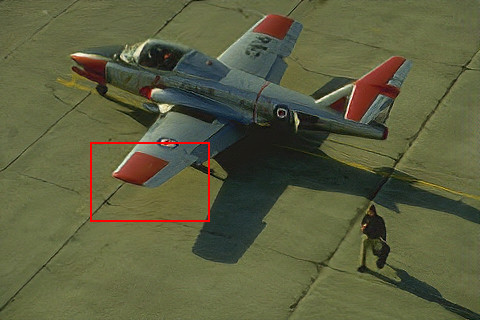}}\smallskip
		\centerline{\includegraphics[width=0.98\linewidth]{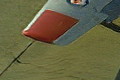}}
	\end{minipage}
	\begin{minipage}[b]{0.24\linewidth}
		\centering
		\centerline{\scriptsize{CX}}\medskip
		\centerline{\includegraphics[width=0.98\linewidth]{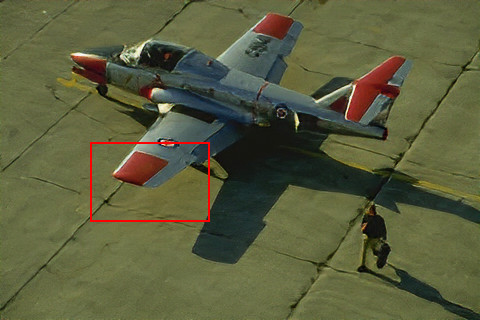}}\smallskip
		\centerline{\includegraphics[width=0.98\linewidth]{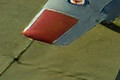}}
	\end{minipage}
	\begin{minipage}[b]{0.24\linewidth}
		\centering
		\centerline{\scriptsize{4PP-EUSR (Ours)}}\medskip
		\centerline{\includegraphics[width=0.98\linewidth]{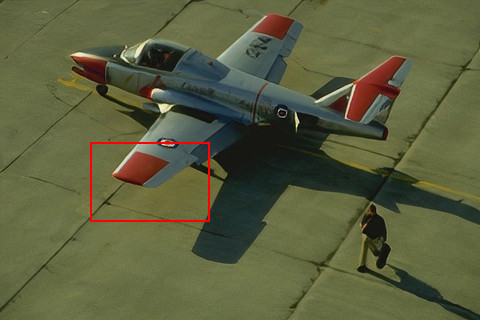}}\smallskip
		\centerline{\includegraphics[width=0.98\linewidth]{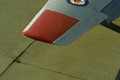}}
	\end{minipage}
	\caption{Images reconstructed by the baselines and our model. The input images are from the BSD100 dataset \cite{martin2001database}.}
	\label{fig:result_baseline_comparison2}
\end{figure*}

\figurename~\ref{fig:result_baseline_comparison} shows example images upscaled by different methods.
Enlarged images of the regions marked by red rectangles are also shown, where high-frequency textures are expected. % as in the ground-truth image.
First, the bicubic interpolation fails to resolve the textures, producing a highly blurred output.
The SRResNet-based, EDSR, MDSR, EUSR, and RCAN models produce richer textures in that region, but still largely blurry.
The output of SRResNet-VGG22 shows distinctive textures, which is due to the employment of a different loss function (i.e., differences of VGG19 features).
Thanks to the adversarial loss, the other models, including SRGANs, CX, and 4PP-EUSR, generate much better outputs in terms of perceptual quality with sacrificing quantitative quality.
Among them, SRGAN-VGG54 and CX recover the most detailed textures, while SRGAN-MSE produces blurry textures.
Our model, 4PP-EUSR, restores the textures more clearly than SRGAN-VGG22 and less distinctly than SRGAN-VGG54.
Nevertheless, ours achieves better quantitative quality than all the SRGANs in terms of PSNR in Table~\ref{table:result_baseline_comparison}.

Another comparison shown in \figurename~\ref{fig:result_baseline_comparison2} further supports the importance of considering both the quantitative and perceptual quality.
Similarly to \figurename~\ref{fig:result_baseline_comparison}, the bicubic interpolation shows the worst output than the others, the models employing only the reconstruction loss (i.e., SRResNets, EDSR, MDSR, EUSR, and RCAN) flatten most of the textured areas, and the rest (i.e., SRGANs, CX, and ours) produce outputs having detailed textures.
However, the SRGAN and CX models tend to exaggerate the indistinct textures on the ground and airplane regions, introducing sizzling artifacts.
For example, the SRGAN-MSE model adds a considerable amount of undesirable noises over the whole image.
On the other hand, thanks to the cooperation of the loss functions, our model successfully recovers much of the textures without any prominent artifacts.

\begin{figure}[t]
	\centering
	\includegraphics[width=0.65\linewidth]{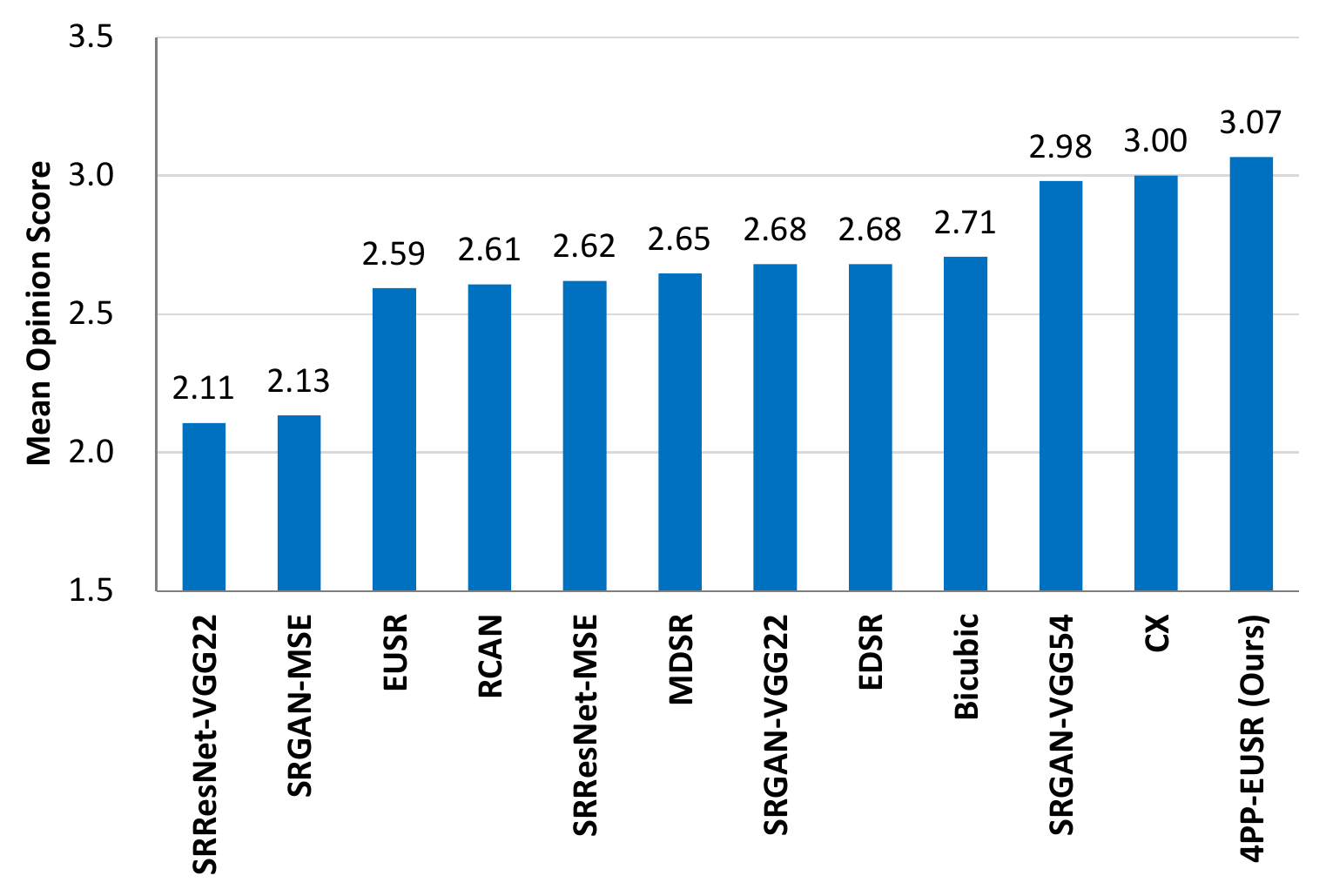}
	\caption{Subjective test results for 10 images of the BSD100 dataset \cite{martin2001database}.}
	\label{fig:subjective_test}
\end{figure}

In addition, we conduct a subjective test to examine the perceptual performance of the super-resolution methods.
We compare the performance of the 12 super-resolution methods for selected ten images in the BSD100 dataset.
We employ 15 participants, which meets the required number of participants for subjective tests in the recommendation ITU-R BT.500-13 \cite{union2012recommendation}.
As for the evaluation method, we follow the same procedure used in \cite{blau20182018}: For a given test image, each participant is asked to rate each of the 120 images on a four-point scale raging among 1 (definitely fake), 2 (probably fake), 3 (probably real), and 4 (definitely real).

\figurename~\ref{fig:subjective_test} summarizes the result of the subjective test.
It demonstrates that our model outperforms the other methods in terms of the mean opinion score.
Our model gets a mean opinion score of 3.07, which means that people regard the output images of ours as ``probably real'' ones.
SRGAN-MSE and SRResNet-VGG22 get the lowest opinion scores among the compared methods.
As shown in \figurename~\ref{fig:result_baseline_comparison2}, it is due to the excessive amount of undesirable artifact introduced in the super-resolved images.
The result supports that considering both quantitative and perceptual quality in our model is helpful to obtain visually pleasant upscaled images.

\subsection{Comparing upscaling paths}
\label{sec:comparing_upscaling_paths}

As described in Section~\ref{sec:multipass_upscaling} and shown in \figurename~\ref{fig:4pp_eusr_structure}, our model produces three upscaled images by utilizing all the upscaling paths: by passing through the $\times$4 path, by passing two times through the $\times$2 path, and by passing through the $\times$8 path and then downscaling via bicubic interpolation.
Here, we compare the results obtained from the different upscaling paths to examine what aspects our model considers to learn.

\begin{figure*}[t]
	\centering
	\begin{minipage}[b]{0.24\linewidth}
		\centering
		\centerline{\scriptsize{Ground-truth}}\medskip
		\centerline{\includegraphics[width=0.98\linewidth]{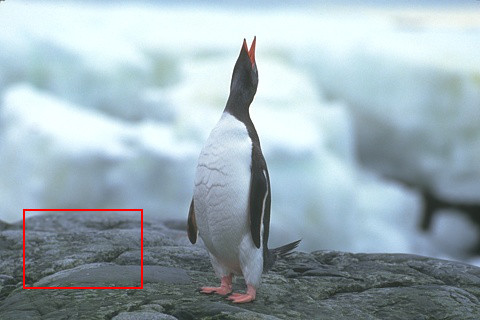}}\smallskip
		\centerline{\includegraphics[width=0.98\linewidth]{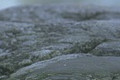}}
	\end{minipage}
	\begin{minipage}[b]{0.24\linewidth}
		\centering
		\centerline{\scriptsize{$\times$4 path}}\medskip
		\centerline{\includegraphics[width=0.98\linewidth]{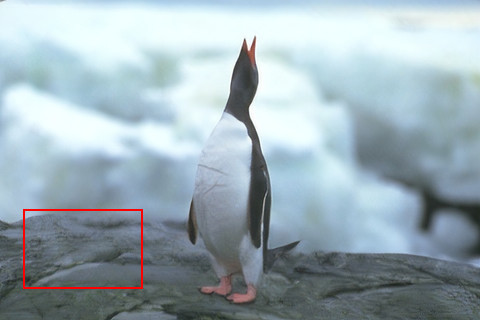}}\smallskip
		\centerline{\includegraphics[width=0.98\linewidth]{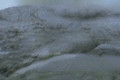}}
	\end{minipage}
	\begin{minipage}[b]{0.24\linewidth}
		\centering
		\centerline{\scriptsize{$\times$2 path -- $\times$2 path}}\medskip
		\centerline{\includegraphics[width=0.98\linewidth]{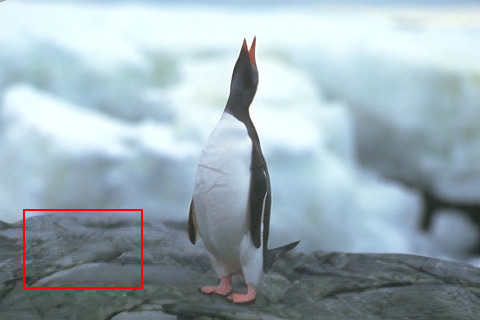}}\smallskip
		\centerline{\includegraphics[width=0.98\linewidth]{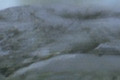}}
	\end{minipage}
	\begin{minipage}[b]{0.24\linewidth}
		\centering
		\centerline{\scriptsize{$\times$8 path -- downscale}}\medskip
		\centerline{\includegraphics[width=0.98\linewidth]{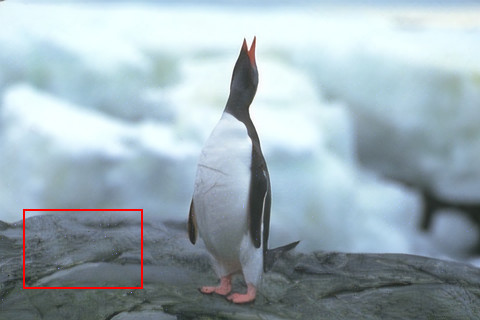}}\smallskip
		\centerline{\includegraphics[width=0.98\linewidth]{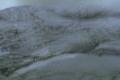}}
	\end{minipage}
	\caption{Images reconstructed by different upscaling paths of our model. The input and ground-truth images are from the BSD100 dataset \cite{martin2001database}.}
	\label{fig:result_upscale_path_comparison}
\end{figure*}

\begin{table}[t]
	\setlength{\tabcolsep}{0.4em}
	\scriptsize
	\centering
	\caption{Performance comparison of the outputs obtained from different three upscaling paths of the 4PP-EUSR model. The results are for the Set5 \cite{bevilacqua2012low}, Set14 \cite{zeyde2010single}, and BSD100 \cite{martin2001database} datasets.}
	\label{table:result_upscale_path_comparison}
	\begin{tabular}{lccccc}
		\textbf{Set5} & PSNR (dB) & SSIM & NIQE & SR score & PI \\
		\noalign{\smallskip}
		\hline
		\noalign{\smallskip}
		$\times$4 & 31.369 & 0.870 & 5.366 & 6.890 & 4.238 \\
		$\times$2 -- $\times$2 & 31.491 & 0.875 & 6.500 & 6.887 & 4.806 \\
		$\times$8 -- downscale & 31.255 & 0.867 & 6.044 & 7.008 & 4.518 \\
		\noalign{\smallskip}
		\noalign{\smallskip}
		\textbf{Set14} & PSNR (dB) & SSIM & NIQE & SR score & PI \\
		\noalign{\smallskip}
		\hline
		\noalign{\smallskip}
		$\times$4 & 27.969 & 0.751 & 4.147 & 7.457 & 3.345 \\
		$\times$2 -- $\times$2 & 28.096 & 0.759 & 4.858 & 7.429 & 3.714 \\
		$\times$8 -- downscale & 27.906 & 0.750 & 4.631 & 7.684 & 3.474 \\
		\noalign{\smallskip}
		\noalign{\smallskip}
		\textbf{BSD100} & PSNR (dB) & SSIM & NIQE & SR score & PI \\
		\noalign{\smallskip}
		\hline
		\noalign{\smallskip}
		$\times$4 & 26.904 & 0.701 & 3.820 & 7.907 & 2.956 \\
		$\times$2 -- $\times$2 & 27.080 & 0.710 & 4.951 & 7.812 & 3.570 \\
		$\times$8 -- downscale & 26.844 & 0.699 & 4.584 & 8.156 & 3.214
	\end{tabular}
\end{table}

Table~\ref{table:result_upscale_path_comparison} compares the performance of the three upscaling paths of our model.
While the PSNR and SSIM values are very similar among the three cases, the $\times$4 path shows the best performance in terms of the NIQE and PI values.
This implies that upscaling using the $\times$2 path or $\times$8 path is more difficult than the $\times$4 path.

\figurename~\ref{fig:result_upscale_path_comparison} shows an example result showing large differences between the three cases.
The appearances of the textures in the enlarged regions are different depending on the upscaling paths, although the overall patterns of the textures follow that of the ground-truth image.
First, the output obtained by the two-pass upscaling using the $\times$2 path contains grid-like textures.
One possible reason is due to the uncertainty in the order of passing: the model does not know whether the current input image is firstly or secondly inputted between the two passes, thus the two-pass upscaling is not fully optimized.
Second, the output obtained from the $\times$8 path with downscaling has unexpected white and black pixels, which are similar to the salt-and-pepper noise.
It seems that since such noise tends to be removed by downscaling, inclusion of the noise in the output is not necessarily avoided during the training of the $\times$8 path.
These results show that each upscaling path of our model learns a different strategy for super-resolution and thus the model is trained to cope with various types of textures via the shared part of the upscaling paths (i.e., the intermediate residual module shown in \figurename~\ref{fig:eusr_structure}).

\subsection{Effectiveness of multi-pass upscaling}
\label{sec:multipass_effectiveness}

\begin{table}[t]
	\setlength{\tabcolsep}{0.4em}
	\scriptsize
	\centering
	\caption{Performance comparison of the 4PP-EUSR models trained with and without multi-pass upscaling for the Set5 \cite{bevilacqua2012low}, Set14 \cite{zeyde2010single}, and BSD100 \cite{martin2001database} datasets.}
	\label{table:result_multipass_comparison}
	\begin{tabular}{lccccc}
		\textbf{Set5} & PSNR (dB) & SSIM & NIQE & SR score & PI \\
		\noalign{\smallskip}
		\hline
		\noalign{\smallskip}
		With multi-pass & 31.369 & 0.870 & 5.366 & 6.890 & 4.238 \\
		Without multi-pass & 31.320 & 0.869 & 5.917 & 6.835 & 4.541 \\
		\noalign{\smallskip}
		\noalign{\smallskip}
		\textbf{Set14} & PSNR (dB) & SSIM & NIQE & SR score & PI \\
		\noalign{\smallskip}
		\hline
		\noalign{\smallskip}
		With multi-pass & 27.969 & 0.751 & 4.147 & 7.457 & 3.345 \\
		Without multi-pass & 27.699 & 0.742 & 4.221 & 7.594 & 3.313 \\
		\noalign{\smallskip}
		\noalign{\smallskip}
		\textbf{BSD100} & PSNR (dB) & SSIM & NIQE & SR score & PI \\
		\noalign{\smallskip}
		\hline
		\noalign{\smallskip}
		With multi-pass & 26.904 & 0.701 & 3.820 & 7.907 & 2.956 \\
		Without multi-pass & 26.614 & 0.688 & 4.327 & 8.140 & 3.093 \\
	\end{tabular}
\end{table}

The 4PP-EUSR model employs multi-pass upscaling as aforementioned in Section~\ref{sec:multipass_upscaling}.
To investigate its effectiveness, we compare the performance of the models trained with and without multi-pass upscaling.

Table~\ref{table:result_multipass_comparison} shows the performance measures of the models in terms of the PSNR, SSIM, NIQE, SR score, and PI values.
It demonstrates that employing multi-pass upscaling is beneficial to enhance both the quantitative and perceptual quality.
The model trained with multi-pass upscaling shows larger PSNR and SSIM values and smaller NIQE values for all the three datasets, and smaller PI values for the datasets except Set14.
This confirms that the multi-pass upscaling can improve the overall quality of the upscaled images.

\subsection{Roles of loss functions}
\label{sec:roles_of_loss_functions}

\begin{figure*}[t]
	\centering
	\begin{minipage}[b]{0.16\linewidth}
		\centering
		\centerline{\scriptsize{Ground-truth}}\medskip
		\centerline{\includegraphics[width=0.98\linewidth]{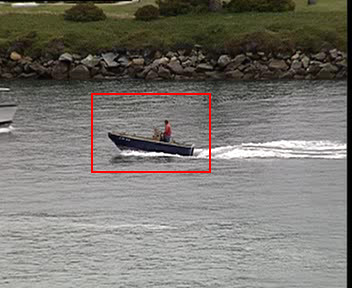}}\smallskip
		\centerline{\includegraphics[width=0.98\linewidth]{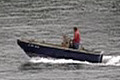}}
	\end{minipage}
	\begin{minipage}[b]{0.16\linewidth}
		\centering
		\centerline{\scriptsize{With all losses}}\medskip
		\centerline{\includegraphics[width=0.98\linewidth]{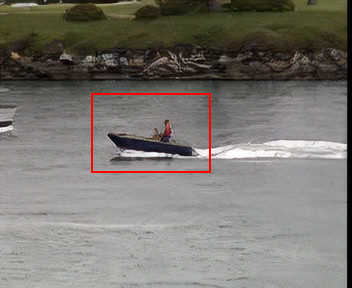}}\smallskip
		\centerline{\includegraphics[width=0.98\linewidth]{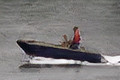}}
	\end{minipage}
	\begin{minipage}[b]{0.16\linewidth}
		\centering
		\centerline{\scriptsize{Without ${l}_{r}$}}\medskip
		\centerline{\includegraphics[width=0.98\linewidth]{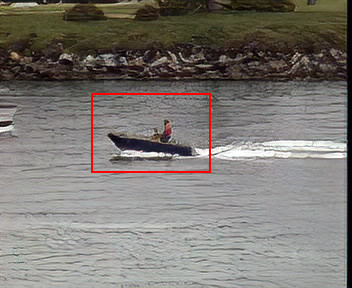}}\smallskip
		\centerline{\includegraphics[width=0.98\linewidth]{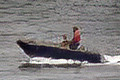}}
	\end{minipage}
	\begin{minipage}[b]{0.16\linewidth}
		\centering
		\centerline{\scriptsize{Without ${l}_{g}$}}\medskip
		\centerline{\includegraphics[width=0.98\linewidth]{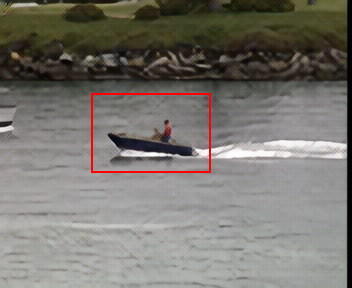}}\smallskip
		\centerline{\includegraphics[width=0.98\linewidth]{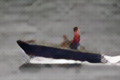}}
	\end{minipage}
	\begin{minipage}[b]{0.16\linewidth}
		\centering
		\centerline{\scriptsize{Without ${l}_{as}$, ${l}_{ar}$}}\medskip
		\centerline{\includegraphics[width=0.98\linewidth]{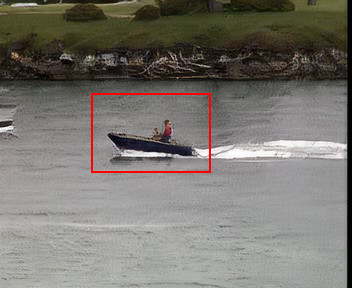}}\smallskip
		\centerline{\includegraphics[width=0.98\linewidth]{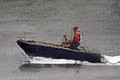}}
	\end{minipage}
	\begin{minipage}[b]{0.16\linewidth}
		\centering
		\centerline{\scriptsize{Without ${l}_{ss}$, ${l}_{sr}$}}\medskip
		\centerline{\includegraphics[width=0.98\linewidth]{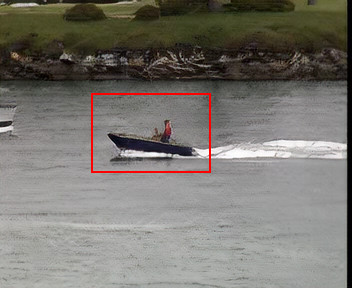}}\smallskip
		\centerline{\includegraphics[width=0.98\linewidth]{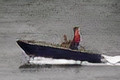}}
	\end{minipage}
	\caption{Images reconstructed by our models trained with excluding specific loss functions. The input and ground-truth images are from the Set14 dataset \cite{zeyde2010single}.}
	\label{fig:result_loss_comparison}
\end{figure*}

\begin{table}[t]
	\setlength{\tabcolsep}{0.4em}
	\scriptsize
	\centering
	\caption{Performance comparison of the 4PP-EUSR models trained by excluding specific loss functions. The models are evaluated on the Set5 \cite{bevilacqua2012low}, Set14 \cite{zeyde2010single}, and BSD100 \cite{martin2001database} datasets.}
	\label{table:result_loss_comparison}
	\begin{tabular}{lccccc}
		\textbf{Set5} & PSNR (dB) & SSIM & NIQE & SR score & PI \\
		\noalign{\smallskip}
		\hline
		\noalign{\smallskip}
		With all losses & 31.369 & 0.870 & 5.366 & 6.890 & 4.238 \\
		Without ${l}_{r}$ & 29.252 & 0.834 & 5.121 & 8.434 & 3.344 \\
		Without ${l}_{g}$ & 32.145 & 0.891 & 6.665 & 5.687 & 5.489 \\
		Without ${l}_{as}$, ${l}_{ar}$ & 30.974 & 0.862 & 5.503 & 7.432 & 4.035 \\
		Without ${l}_{ss}$, ${l}_{sr}$ & 31.389 & 0.873 & 5.406 & 6.807 & 4.300 \\
		\noalign{\smallskip}
		\noalign{\smallskip}
		\textbf{Set14} & PSNR (dB) & SSIM & NIQE & SR score & PI \\
		\noalign{\smallskip}
		\hline
		\noalign{\smallskip}
		With all losses & 27.969 & 0.751 & 4.147 & 7.457 & 3.345 \\
		Without ${l}_{r}$ & 26.137 & 0.705 & 4.187 & 8.132 & 3.028 \\
		Without ${l}_{g}$ & 28.589 & 0.779 & 5.287 & 6.153 & 4.567 \\
		Without ${l}_{as}$, ${l}_{ar}$ & 27.601 & 0.738 & 3.976 & 7.804 & 3.086 \\
		Without ${l}_{ss}$, ${l}_{sr}$ & 27.853 & 0.752 & 4.026 & 7.571 & 3.228 \\
		\noalign{\smallskip}
		\noalign{\smallskip}
		\textbf{BSD100} & PSNR (dB) & SSIM & NIQE & SR score & PI \\
		\noalign{\smallskip}
		\hline
		\noalign{\smallskip}
		With all losses & 26.904 & 0.701 & 3.820 & 7.907 & 2.956 \\
		Without ${l}_{r}$ & 25.142 & 0.649 & 4.118 & 8.773 & 2.673 \\
		Without ${l}_{g}$ & 27.546 & 0.734 & 5.362 & 6.403 & 4.480 \\
		Without ${l}_{as}$, ${l}_{ar}$ & 26.540 & 0.684 & 4.016 & 8.343 & 2.837 \\
		Without ${l}_{ss}$, ${l}_{sr}$ & 26.870 & 0.703 & 3.780 & 7.989 & 2.895
	\end{tabular}
\end{table}

Our model employs multiple types of loss functions as described in Section~\ref{sec:training_perceptual_sr}.
To analyze the role of each loss function, we conduct an experiment where our model is trained with excluding specific loss functions.
In detail, we obtain the models trained without ${l}_{r}$, without ${l}_{g}$, without ${l}_{as}$ and ${l}_{ar}$, and without ${l}_{ss}$ and ${l}_{sr}$.

Table~\ref{table:result_loss_comparison} shows the PSNR, SSIM, NIQE, SR score, and PI values of the trained models.
First, excluding ${l}_{r}$ deteriorates the quantitative quality of the upscaled images, showing smaller PSNR values, and improves the perceptual quality, showing smaller PI values, in comparison to the model trained with all losses.
Excluding ${l}_{g}$ results in the opposite outcomes: it increases the quantitative quality (i.e., larger PSNR values) and decreases the perceptual quality (i.e., larger PI values).
Excluding the aesthetic losses (i.e., ${l}_{as}$ and ${l}_{ar}$) or subjective losses (i.e., ${l}_{ss}$ and ${l}_{sr}$) also affects to the performance in terms of PSNR.

\figurename~\ref{fig:result_loss_comparison} shows example output images, where more evident differences of the roles of the loss functions can be observed.
First, the image obtained from the model trained without the reconstruction loss (i.e., ${l}_{r}$) contains the most distinct textures than the others, but the overall color distribution is slightly different from that of the ground-truth image.
On the other hand, the result generated by the model trained without the adversarial loss (i.e., ${l}_{g}$) preserves the overall structure of the ground-truth image, while its details are more blurry than those of the others.
The output of the model trained without the subjective loss functions contains more lattice-like textures than that of the model trained without the aesthetic loss functions.
This implies that the aesthetic losses contribute to the restoration of highly structured textures, while the subjective losses are helpful to construct dispersed high-frequency textures.
Finally, the image obtained by training with all the proposed loss functions is the most reliable and natural.

\subsection{Comparing different loss weights}
\label{sec:comparing_different_loss_weights}

\begin{table}[t]
	\setlength{\tabcolsep}{0.4em}
	\scriptsize
	\centering
	\caption{Performance comparison of our models trained with different combinations of the loss weights. The models are evaluated on the Set5 \cite{bevilacqua2012low}, Set14 \cite{zeyde2010single}, and BSD100 \cite{martin2001database} datasets.}
	\label{table:result_loss_weight_comparison}
	\begin{tabular}{lccccc}
		\textbf{Set5} & PSNR (dB) & SSIM & NIQE & SR score & PI \\
		\noalign{\smallskip}
		\hline
		\noalign{\smallskip}
		${\alpha}_{p}$= 0, ${\alpha}_{r}$= 0.5 & 31.891 & 0.881 & 6.386 & 5.637 & 5.375 \\
		${\alpha}_{p}$= 0, ${\alpha}_{r}$= 0.05 & 31.748 & 0.880 & 5.739 & 6.073 & 4.833 \\
		${\alpha}_{p}$= 0, ${\alpha}_{r}$= 0.005 & 30.504 & 0.854 & 5.730 & 7.633 & 4.048 \\
		${\alpha}_{p}$= 1, ${\alpha}_{r}$= 0.5 & 31.839 & 0.881 & 6.242 & 5.956 & 5.143 \\
		${\alpha}_{p}$= 1, ${\alpha}_{r}$= 0.05 & 31.369 & 0.870 & 5.366 & 6.890 & 4.238 \\
		${\alpha}_{p}$= 1, ${\alpha}_{r}$= 0.005 & 30.753 & 0.857 & 5.234 & 7.685 & 3.775 \\
		\noalign{\smallskip}
		\noalign{\smallskip}
		\textbf{Set14} & PSNR (dB) & SSIM & NIQE & SR score & PI \\
		\noalign{\smallskip}
		\hline
		\noalign{\smallskip}
		${\alpha}_{p}$= 0, ${\alpha}_{r}$= 0.5 & 28.348 & 0.765 & 4.597 & 6.852 & 3.872 \\
		${\alpha}_{p}$= 0, ${\alpha}_{r}$= 0.05 & 28.218 & 0.764 & 4.154 & 7.099 & 3.527 \\
		${\alpha}_{p}$= 0, ${\alpha}_{r}$= 0.005 & 26.864 & 0.715 & 4.644 & 7.897 & 3.374 \\
		${\alpha}_{p}$= 1, ${\alpha}_{r}$= 0.5 & 28.316 & 0.763 & 4.766 & 6.843 & 3.961 \\
		${\alpha}_{p}$= 1, ${\alpha}_{r}$= 0.05 & 27.969 & 0.751 & 4.147 & 7.457 & 3.345 \\
		${\alpha}_{p}$= 1, ${\alpha}_{r}$= 0.005 & 27.020 & 0.726 & 4.017 & 7.970 & 3.023 \\
		\noalign{\smallskip}
		\noalign{\smallskip}
		\textbf{BSD100} & PSNR (dB) & SSIM & NIQE & SR score & PI \\
		\noalign{\smallskip}
		\hline
		\noalign{\smallskip}
		${\alpha}_{p}$= 0, ${\alpha}_{r}$= 0.5 & 27.332 & 0.717 & 4.633 & 6.932 & 3.850 \\
		${\alpha}_{p}$= 0, ${\alpha}_{r}$= 0.05 & 27.162 & 0.715 & 3.987 & 7.478 & 3.254 \\
		${\alpha}_{p}$= 0, ${\alpha}_{r}$= 0.005 & 25.833 & 0.659 & 5.374 & 8.548 & 3.413 \\
		${\alpha}_{p}$= 1, ${\alpha}_{r}$= 0.5 & 27.271 & 0.714 & 4.498 & 7.042 & 3.728 \\
		${\alpha}_{p}$= 1, ${\alpha}_{r}$= 0.05 & 26.904 & 0.701 & 3.820 & 7.907 & 2.956 \\
		${\alpha}_{p}$= 1, ${\alpha}_{r}$= 0.005 & 26.176 & 0.678 & 3.867 & 8.552 & 2.657
	\end{tabular}
\end{table}

\begin{figure}[t]
	\centering
	\begin{minipage}[b]{0.24\linewidth}
		\centering
		\centerline{\scriptsize{Ground-truth}}\medskip
		\centerline{\includegraphics[width=0.98\linewidth]{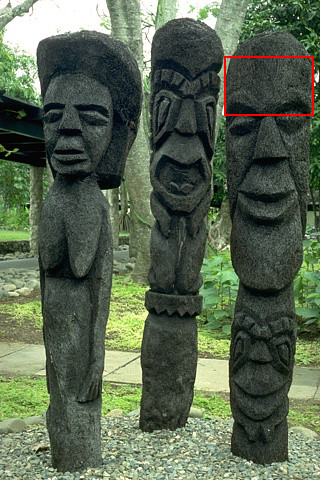}}\smallskip
		\centerline{\includegraphics[width=0.98\linewidth]{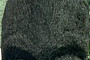}}
	\end{minipage}
	\begin{minipage}[b]{0.24\linewidth}
		\centering
		\centerline{\scriptsize{${\alpha}_{p} = 1, {\alpha}_{r} = 0.5$}}\medskip
		\centerline{\includegraphics[width=0.98\linewidth]{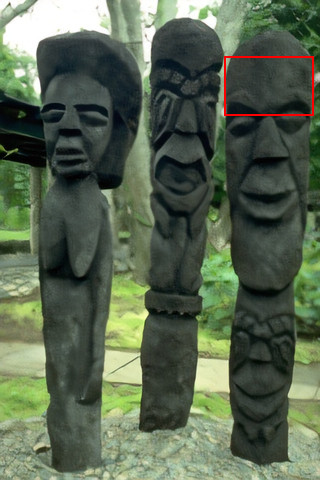}}\smallskip
		\centerline{\includegraphics[width=0.98\linewidth]{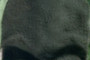}}
	\end{minipage}
	\begin{minipage}[b]{0.24\linewidth}
		\centering
		\centerline{\scriptsize{${\alpha}_{p} = 1, {\alpha}_{r} = 0.05$}}\medskip
		\centerline{\includegraphics[width=0.98\linewidth]{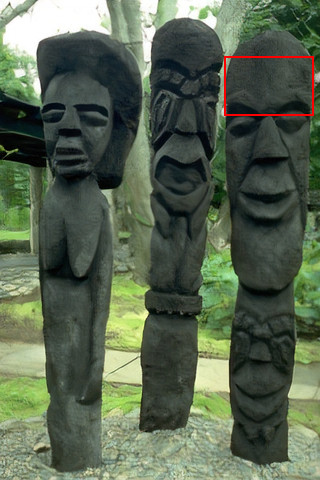}}\smallskip
		\centerline{\includegraphics[width=0.98\linewidth]{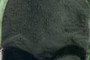}}
	\end{minipage}
	\begin{minipage}[b]{0.24\linewidth}
		\centering
		\centerline{\scriptsize{${\alpha}_{p} = 1, {\alpha}_{r} = 0.005$}}\medskip
		\centerline{\includegraphics[width=0.98\linewidth]{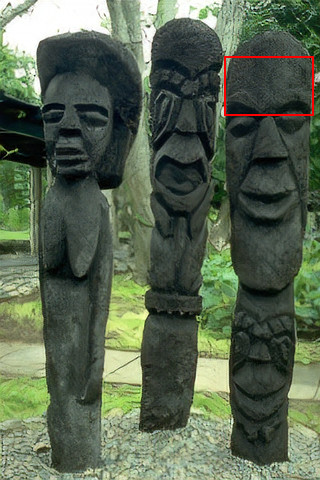}}\smallskip
		\centerline{\includegraphics[width=0.98\linewidth]{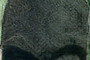}}
	\end{minipage}
	\caption{Images reconstructed by our models trained with different combinations of the loss weights. The input and ground-truth images are from the BSD100 dataset \cite{martin2001database}.}
	\label{fig:result_loss_weight_comparison}
\end{figure}

Finally, we train our model with different weights of the loss functions.
Specifically, we alter the weight of the reconstruction loss in (\ref{eq:default_loss_equation}) as
\begin{equation}
\label{eq:loss_equation_variable_lr}
l = {\alpha}_{r} {l}_{r} + 0.1 {l}_{g} + {\alpha}_{p} ( 0.01 {l}_{as} + 0.1 {l}_{ar} + 0.01 {l}_{ss} + 0.1 {l}_{sr} )
\end{equation}
with ${\alpha}_{r} \in \{0.5, 0.05, 0.005\}$ and ${\alpha}_{p} \in \{0, 1\}$.
We can expect that a larger ${\alpha}_{r}$ value leads the model to be trained towards producing outputs having better quantitative quality.
The term ${\alpha}_{p}$ determines whether to use the score predictors or not.
%Note that decreasing ${\alpha}_{r}$ not only reduces the contribution of reconstruction loss ${l}_{r}$ but also increases the contribution of adversarial loss ${l}_{g}$.

Table~\ref{table:result_loss_weight_comparison} presents the performance of our model trained with different weight values.
As expected, decreasing the level of contribution of the reconstruction loss with a smaller ${\alpha}_{r}$ results in lower PSNR values.
On the other hand, the PI values are also decreased, which indicates improved qualitative quality.
These observations emerge as the visual differences of the upscaled images shown in \figurename~\ref{fig:result_loss_weight_comparison}.
When we examine the enlarged regions where high-frequency textures are expected, a decreased ${\alpha}_{r}$ value affects the clearness of the output images, due to relatively larger contributions of the adversarial and perceptual losses.
These confirm that there is a tradeoff between quantitative and perceptual quality as mentioned in \cite{blau2017perception}, and our model has a capability to deal with the priorities of these quality measures by adjusting the weights of the loss functions.
In addition, the result shows that employing the score predictors (i.e., with ${\alpha}_{p}=1$) is helpful to improve the perceptual quality of the upscaled images, which can be observed as decreased PI values in Table~\ref{table:result_loss_weight_comparison}.

\section{Conclusion}
\label{sec:conclusion}

In this paper, we proposed a perceptually improved super-resolution method, which employs multi-pass image restoration via a multi-scale super-resolution model and trains the model with a discriminator network and two qualitative score predictors.
The results showed that our model successfully recovers the original textures in a perceptual manner while preventing quantitative quality degradation.

\section*{Acknowledgements}

This research was supported by the MSIT (Ministry of Science and ICT), Korea, under the ``ICT Consilience Creative Program'' (IITP-2018-2017-0-01015) supervised by the IITP (Institute for Information \& communications Technology Promotion). In addition, this work was also supported by the IITP grant funded by the Korea government (MSIT) (R7124-16-0004, Development of Intelligent Interaction Technology Based on Context Awareness and Human Intention Understanding).

\bibliographystyle{splncs04}
\bibliography{arxiv}
\end{document}